\documentclass[final,5p,times,twocolumn]{elsarticle}
\usepackage[T1]{fontenc}
\usepackage[english]{babel}
\usepackage{textcomp}
\usepackage[table, xcdraw, pdftex, usenames, dvipsnames]{xcolor}
\usepackage{xspace}
\usepackage{graphicx}
\usepackage{amsfonts}
\usepackage{amsmath}
\usepackage{amsthm}
\usepackage{booktabs}
\usepackage[acronym]{glossaries}
\usepackage[normalem]{ulem}
\usepackage[hidelinks, bookmarksopen=true]{hyperref}
\hypersetup{
    colorlinks=true,
    allcolors=blue,
    pdftitle={Process mining-driven modeling and simulation to enhance fault diagnosis in cyber-physical systems},
    pdfpagemode=FullScreen,
}
\usepackage[capitalize,noabbrev]{cleveref}
\usepackage{tikz}
\usepackage{pgf}
\usepackage{pgfplots}
\usepackage{subfigure}
\usepackage{multirow}
\usepackage{colortbl}
\usepackage{siunitx}
\usepackage{algorithm}
\usepackage{algpseudocode}
\usepackage{diagbox}
\usepackage{robotarm}
\usepackage{epstopdf}
\usepackage{comment}
\usepackage{nicefrac}
\usepackage{colortbl}

\newtheorem{definition}{Definition}[section]

\usetikzlibrary{fit}
\usetikzlibrary{math}
\usetikzlibrary{calc}
\usetikzlibrary{petri}
\usetikzlibrary{arrows}
\usetikzlibrary{arrows.meta}
\usetikzlibrary{shapes.arrows}
\usetikzlibrary{shapes.symbols}
\usetikzlibrary{shapes.geometric}
\usetikzlibrary{patterns}
\usetikzlibrary{patterns.meta}
\usetikzlibrary{positioning}
\pgfplotsset{compat=newest}
\makeatletter
\tikzset{%
    >=stealth,
    ultra thin/.style= {line width=0.1pt},
    very thin/.style=  {line width=0.2pt},
    thin/.style=       {line width=0.4pt},% thin is the default
    semithick/.style=  {line width=0.6pt},
    thick/.style=      {line width=0.8pt},
    very thick/.style= {line width=1.2pt},
    ultra thick/.style={line width=1.6pt},
    fblue/.style={fill=blue!50},
    fred/.style={fill=red!50},
    forange/.style={fill=orange!50},
    fgreen/.style={fill=green!50},
    fgray/.style={fill=gray!50},
    lblue/.style={draw=blue},
    lred/.style={draw=red},
    lorange/.style={draw=orange},
    lgreen/.style={draw=green},
    lgray/.style={draw=gray},
    AN/.style={anchor=north},
    ANW/.style={anchor=north west},
    ANE/.style={anchor=north east},
    AE/.style={anchor=east},
    AW/.style={anchor=west},
    AS/.style={anchor=south},
    ASW/.style={anchor=south west},
    ASE/.style={anchor=south east},
    AC/.style={anchor=center},
	opaque node/.code 2 args={\tikzset{opacity=#1, text opacity=#2}},
	double color fill/.code 2 args={%
		\pgfdeclareverticalshading[%
		tikz@axis@top,tikz@axis@middle,tikz@axis@bottom%
		]{diagonalfill}{100bp}{%
			color(0bp)=(tikz@axis@bottom);%
			color(50bp)=(tikz@axis@bottom);%
			color(50bp)=(tikz@axis@middle);%
			color(50bp)=(tikz@axis@top);%
			color(100bp)=(tikz@axis@top)%
		}%
		\tikzset{%
			shade,%
			left color=#1,%
			right color=#2,%
			shading=diagonalfill%
		}%
	}%
}
\makeatother

\definecolor{red}    {HTML}{b7211f}
\definecolor{green}  {HTML}{147546}
\definecolor{orange}           {HTML}{FFA500}
\definecolor{plum}             {HTML}{DDA0DD}
\definecolor{purple}           {HTML}{9370DB}
\definecolor{purple_protege}   {HTML}{bf23ba}
\definecolor{blue_protege}     {HTML}{2fb9c0}
\definecolor{darkblue_protege} {HTML}{0a5ea8}

\definecolor{blockbg}{HTML} {E7EEF8}

\definecolor{background}{HTML} {F2F2F2}
\definecolor{keyword}{HTML}    {426B8C}
\definecolor{string}{HTML}     {338535}
\definecolor{comment}{HTML}    {0F52BA}

\usepackage{etoolbox}    % For \iftoggle
\usepackage{xcolor}      % For \color
\usepackage[normalem]{ulem} % For \sout and \uline
\usepackage{ifthen}      % Optional for extra logic

\newtoggle{ShowChanges}
\toggletrue{ShowChanges} % Or \togglefalse

\usepackage{glossaries}

\newacronym{aadl}{AADL}{Architecture Analysis \& Design Language}
\newacronym{ac}{AC}{Automation Controller}
\newacronym{adc}{ADC}{Analog-to-Digital Converter}
\newacronym{ag}{A/G}{Assume-Guarantee}
\newacronym{amba}{AMBA}{Advanced Microcontroller Bus Architecture}
\newacronym{ams}{AMS}{Analog-Mixed Signals}
\newacronym{api}{API}{Application Programming Interface}
\newacronym{aml}{AutomationML}{Automation Markup Language}
\newacronym{amqp}{AMQP}{Advanced Message Queuing Protocol}
\newacronym{agv}{AGV}{Automated Guided Vehicles}

\newacronym{bom}{BOM}{Bill of Materials}
\newacronym{bdd}{BDD}{Block Definition Diagram}
\newacronym{bpmn}{BPMN}{Business Process Modeling Notation}

\newacronym{csp}{CSP}{Constraint Satisfaction Programming}
\newacronym{cad}{CAD}{Computer-aided design}
\newacronym{cnc}{CNC}{Computer Numerical Control}
\newacronym{casse}{CASSE}{Communication Aware Specification and Synthesis
	Environment}
\newacronym{cps}{CPS}{Cyber-Physical System}
\newacronym{cpps}{CPPS}{Cyber-Physical Production System}

\newacronym{caex}{CAEX}{Computer Aided Engineering Exchange}
\newacronym{collada}{COLLADA}{COLLAborative Design Activity}
\newacronym{cc}{CC}{Conformance Checking}

\newacronym{dse}{DSE}{Design Space Exploration}
\newacronym{dac}{DAC}{Digital-to-Analog Converter}
\newacronym{dsl}{DSL}{Domain Specification Language}
\newacronym{din}{DIN}{Deutsches Institut f\"ur Normung}
\newacronym{b2mml}{B2MML}{Business To Manufacturing Markup Language}
\newacronym{dfjss}{DFJSS}{Dynamic Flexible Job Shop Scheduling}

\newacronym{eda}{EDA}{Electronic Design Automation}
\newacronym[longplural=Electromagnetic Interferences]{emi}{EMI}{Electromagnetic Interference}
\newacronym{esl}{ESL}{Electronic System Level}
\newacronym{eln}{ELN}{Electrical Linear Network}
\newacronym{erp}{ERP}{Enterprise Resource Planning}

\newacronym{fmi}{FMI}{Functional Mock-up Interface}
\newacronym{fms}{FMS}{Flexible Manufacturing System}
\newacronym{fmu}{FMU}{Functional Mock-up Unit}
\newacronym{fmea}{FMEA}{Failure Mode and Effect Analysis}
\newacronym{fsm}{FSM}{Finite State Machine}
\newacronym{fdm}{FDM}{Fused Deposition Modeling}
\newacronym{fjss}{FJSS}{Flexible Job Shop Scheduling}
\newacronym{fem}{FEM}{Finite Element Model}
\newacronym{f}{F}{Fitness}

\newacronym{gr1}{GR(1)}{General Reactivity}
\newacronym{gmm}{GMM}{Gaussian Mixture Model}

\newacronym{hdl}{HDL}{Hardware Description Language}
\newacronym{hif}{HIF}{Heterogeneous Intermediate Format}
\newacronym{hrm}{HRM}{Hardware Resource Modeling}
\newacronym{hw}{HW}{Hardware}
\newacronym{hmi}{HMI}{Human-Machine Interaction}

\newacronym{ic}{IC}{Integrated Circuit}
\newacronym{ice}{ICE}{Industrial Computer Engineering}
\newacronym{ios}{IoS}{Internet of Services}
\newacronym{iot}{IoT}{Internet of Things}
\newacronym{iiot}{IIoT}{Industrial Internet of Things}
\newacronym{isa}{ISA}{International Society of Automation}
\newacronym{iss}{ISS}{Instruction Set Simulator}
\newacronym[longplural={Intellectual Properties}]{ip}{IP}{Intellectual Property}
\newacronym{ibd}{IBD}{Internal Block Diagram}
\newacronym{im}{IM}{Inductive Miner}
\newacronym{imf}{IMf}{Inductive Miner Infrequent}
\newacronym{ilp}{ILP}{Integer Linear Programming-based Miner}

\newacronym{jss}{JSS}{Job Shop Scheduling}

\newacronym{kpn}{KPN}{Khan Process Networks}

\newacronym{ltl}{LTL}{Linear Temporal Logic}
\newacronym{lptn}{LPTN}{Lumped-Parameter Thermal Network}

\newacronym{mbd}{MBD}{Model-based Design}
\newacronym{mbse}{MBSE}{Model-based System Engineering}
\newacronym{mems}{MEMS}{Micro Electro Mechanical Systems}
\newacronym{milp}{MILP}{Mixed Integer Linear Programming}
\newacronym[longplural={Models of Computation}]{moc}{MoC}{Model of Computation}
\newacronym{ml}{ML}{Machine Learning}
\newacronym{mu}{MU}{Mobile Unit}
\newacronym{mes}{MES}{Manufacturing Execution System}
\newacronym{mom}{MOM}{Manufacturing Operations Management}
\newacronym{m2m}{M2M}{Machine to Machine}
\newacronym{mts}{MTS}{Multivariate Time Series}
\newacronym{mapek}{MAPE-K}{Monitor-Analyze-Plan-Execute over a shared Knowledge}

\newacronym{nes}{NES}{Networked Embedded System}
\newacronym{noc}{NoC}{Network on Chip}
\newacronym[longplural={Non-Functional-Properties}]{nfp}{NFP}{Non-Functional-Property}
\newacronym{nc}{NC}{Number of Clusters}

\newacronym{opcua}{OPC~UA}{OPC Unified Architecture}
\newacronym{ostc}{O$S^3$TC}{Open-Source Smart-System Test Case}
\newacronym{ovp}{OVP}{Open Virtual Platform}

\newacronym{pbd}{PBD}{Platform-Based Design}
\newacronym{plc}{PLC}{Programmable Logic Controller}
\newacronym{pm}{PM}{Process Mining}
\newacronym{ppe}{PPE}{Personal Protective Equipment}
\newacronym{pd}{PD}{Process Discovery}
\newacronym{pn}{PN}{Petri Net}
\newacronym{p}{P}{Precision}

\newacronym{qc}{QC}{Quality Checking}

\newacronym{rtl}{RTL}{Register-Transfer Level}
\newacronym{rmse}{RMSE}{Root Mean Square Error}
\newacronym{rpc}{RPC}{Remote Procedure Call}
\newacronym{rtn}{RTN}{Resource Task Network}
\newacronym{road}{RoAD}{Robotic Arm Dataset}
\newacronym{rul}{RUL}{Remaining Useful Life}

\newacronym{sdk}{SDK}{Software Development Kit}
\newacronym{sld}{SLD}{System-Level Design}
\newacronym{soc}{SoC}{System on a Chip}
\newacronym{srm}{SRM}{Software Resource Modeling}
\newacronym{sw}{SW}{Software}
\newacronym{sme}{SME}{Small and Medium Enterprise}
\newacronym{sysml}{SysML}{System Modeling Language}
\newacronym{scnsl}{SCNSL}{SystemC Network Simulation Library}
\newacronym{soa}{SOA}{Service Oriented Architecture}
\newacronym{som}{SOM}{Service Oriented Manufacturing}
\newacronym{scada}{SCADA}{Supervisory Control and Data Acquisition}
\newacronym{stomp}{STOMP}{Streaming Text
	Oriented Messaging Protocol}
\newacronym{stn}{STN}{State Task Network}

\newacronym{tlm}{TLM}{Transaction-Level Modeling}

\newacronym{uml}{UML}{Unified Modeling Language}

\newacronym{vlsi}{VLSI}{Very Large Scale Integration}
\newacronym{vp}{VP}{Virtual Platform}

\newacronym{wsn}{WSN}{Wireless Sensor Networks}

\newacronym{xai}{XAI}{Explainable Artificial Intelligence}
\newacronym{xmi}{XMI}{XML Metadata Interchange}

\makeatletter
\apptocmd\@bibitem{\color{black}\csname keycolor#1\endcsname}{}{\fail}
\apptocmd\@lbibitem{\color{black}\csname keycolor#2\endcsname}{}{\fail}
\newcommand\citecolor[2][red]{\@namedef{keycolor#2}{\color{#1}}}
\makeatother

\journal{Elsevier}

\begin{document}
\begin{frontmatter}
\title{Process mining-driven modeling and simulation to enhance\\ fault diagnosis in cyber-physical systems}

{\author[label1]{Francesco Vitale} 
\author[label2,label3]{Nicola Dall'Ora}
\author[label2]{Sebastiano Gaiardelli}
\author[label2]{Enrico Fraccaroli}
\author[label1]{Nicola Mazzocca}
\author[label2]{Franco Fummi}

\affiliation[label1]{organization={University of Naples Federico II, Department of Electrical Engineering and Information Technology},
            addressline={Via Claudio, 21}, 
            city={Naples},
            postcode={80125},
            country={Italy}}

\affiliation[label2]{organization={University of Verona, Department of Engineering for Innovation Medicine \\(Section of Engineering and Physics)},
            addressline={Strada le Grazie, 15}, 
            city={Verona},
            postcode={37134},
            country={Italy}}

\affiliation[label3]{organization={Guglielmo Marconi University, Department of Engineering Sciences},
            addressline={Via Plinio, 44}, 
            city={Rome},
            postcode={00193},
            country={Italy}}}

\begin{abstract}
Cyber-Physical Systems (CPSs) tightly interconnect digital and physical operations within production environments, enabling real-time monitoring, control, optimization, and autonomous decision-making that directly enhance manufacturing processes and productivity. The inherent complexity of these systems can lead to faults that require robust and interpretable diagnoses to maintain system dependability and operational efficiency. However, manual modeling of faulty behaviors requires extensive domain expertise and cannot leverage the low-level sensor data of the CPS. Furthermore, although powerful, deep learning-based techniques produce black-box diagnostics that lack interpretability, limiting their practical adoption. 
To address these challenges, we set forth a method that performs unsupervised characterization of system states and state transitions from low-level sensor data, uses several process mining techniques to model faults through interpretable stochastic Petri nets, simulates such Petri nets for a comprehensive understanding of system behavior under faulty conditions, and performs Petri net-based fault diagnosis. 
The method begins with detecting collective anomalies involving multiple samples in low-level sensor data. These anomalies are then transformed into structured event logs, enabling the data-driven discovery of interpretable Petri nets through process mining. By enhancing these Petri nets with timing distributions, the approach supports the simulation of faulty behaviors. 
Finally, faults can be diagnosed online by checking collective anomalies with the Petri nets and the corresponding simulations.
The method is applied to the Robotic Arm Dataset (RoAD), a benchmark collected from a robotic arm deployed in a scale-replica smart manufacturing assembly line. The application to RoAD demonstrates the method's effectiveness in modeling, simulating, and classifying faulty behaviors in CPSs. The modeling results demonstrate that our method achieves a satisfactory interpretability-simulation accuracy trade-off with up to 0.676 arc-degree simplicity, 0.395 R$^2$, and 0.088 RMSE. In addition, the fault identification results show that the method achieves an F1 score of up to 98.925\%, while maintaining a low conformance checking time of 0.020 seconds, which competes with other deep learning-based methods.
\medskip
\end{abstract}


\glsresetall

\end{frontmatter}

\section{Introduction}
\label{sec:introduction}
The integration of physical and cyber processes in modern \glspl{cps} offers significant opportunities to enhance service quality in the manufacturing sector, enabling real-time monitoring, control, optimization, and autonomous decision-making, and ensuring a range of characteristics, such as flexibility, scalability, and adaptability~\cite{wan2022contextawarecpps}. However, the resulting complexity of \glspl{cps} calls for adequate means to ensure system reliability despite the presence of vulnerabilities and occurrence of run-time faults~\cite{friedrich2024reliabilityassessment}. Therefore, achieving high reliability involves the prompt detection of anomalies, accurately determining their root causes, and proactively planning recovery actions~\cite{li2025processmonitoringreview}. In addition, simulation frameworks have become indispensable tools to support the development of reliable production systems, accelerate \gls{cps} development, perform what-if analyses, and design effective recovery strategies~\cite{gueuziec2023systemsimulation, tampouratzis2023simulationframework}.

Although many studies have explored anomaly detection in \glspl{cps}, these approaches are mostly limited to characterizing normal behavior and identifying deviations from it. However, anomalies do not automatically translate into faults, which are defined as “adjudged or hypothesized causes” of an error~\cite{avizienis2004taxonomy}. To analyze and identify faults, we adopt a broader perspective from recent literature~\cite{cai2017bnfd, yu2020cnn, li2021abd}, in which diagnosis extends beyond anomaly detection to include identifying and classifying root causes. Within this view, modeling and simulation become essential tools for examining how faults arise and propagate, as well as for planning corrective actions. Accordingly, our work intends to provide structure and insight to support fault diagnosis rather than limiting itself to anomaly detection.

Despite the high demand for interpretable and comprehensive fault diagnosis methods, most approaches still focus on capturing normal patterns in historical sensor data and flag anomalous trends, which, as said, do not provide further information about the type of anomaly that occurred and cannot provide an insightful root cause analysis~\cite{belay2024mtad}. Besides, these approaches are often based on advanced deep learning methods, which, although powerful and extremely accurate, have limited interpretability~\cite{interpreting_recurrent_neural_networks_in_process_mining:Hanga:2020}. On the other hand, other formalisms can be employed to model and simulate faulty behaviors, including Petri nets, widely accepted due to their clear, process-based semantics. These aspects, especially, facilitate interpretability, a quality increasingly critical in trustworthy artificial intelligence applications~\cite{aalst2022pmhb, liu2022trustworthyai, mozafari2023explainablecc, hu2024sensorselectiondiagnosability}. Yet, existing interpretable Petri net-based approaches either require manual, error-prone modeling by domain experts~\cite{mansour2013pnfd,liu2013cpshierarchicalfaulterrorfailuredependencies,nazemzadeh2013pndesfaultmodeling,bourdeau2019energyconsumption} or do not directly target data-driven modeling of faulty \gls{cps} behaviors from low-level sensor data, requiring domain knowledge to supervise the capture of normal behavior into a Petri net and discriminate such behavior from anomalies~\cite{vitale2023pmunsupervisedadiiot, beyel2024cpsbehavioranalysis, vitale2025pmdt}. Hence, a notable gap persists regarding the data-driven modeling of faulty \gls{cps} behaviors through Petri nets directly from low-level sensor data, particularly within frameworks that offer both predictive power and explanatory transparency.

To provide interpretable detection, simulation, and classification of faulty \gls{cps} behavior, we investigate process mining, whose algorithms allow data-driven discovery of Petri nets (process discovery), enhancements of such Petri nets with a time perspective (process enhancement), and comparing new behavior with the Petri nets to evaluate their alignment (conformance checking)~\cite{aalst2022pmhb}. Specifically, we propose a process mining-driven method for fault diagnosis that begins with isolating anomalous CPS behaviors through collective anomaly detection, which allows identifying anomalous behaviors involving multiple sensor data samples. Then, such data is converted into structured event logs unsupervisedly through clustering. The event logs are used to extract interpretable Petri nets through process discovery and to enrich them with actual timings from the collected data via process enhancement. These Petri nets enable detailed stochastic simulations, providing more insights into system behavior and supporting robust fault classification. Finally, conformance checking between online data, the stochastic Petri nets, and their corresponding simulations is performed for fault diagnosis. 

In summary, the novelty of our method involves:
\begin{itemize}
    \item Unsupervised characterization of manufacturing \gls{cps} states, state transitions, state times, and state times distribution from low-level sensor data;
    \item Integration of process mining to build interpretable Petri nets based on the control-flow relations of state transitions, enhanced with the time perspective of the state times distributions;
    \item Simulating the stochastic Petri nets to have a comprehensive understanding of the states traversed by the system and their timings under faulty conditions;
    \item Petri net-based fault diagnosis by isolating anomalous behavior and checking its alignment to the faults characterized offline, allowing the interpretable identification of different fault types.
\end{itemize}

We demonstrate our method's capabilities in modeling accuracy, simulation fidelity, and fault diagnosis effectiveness in smart manufacturing, referencing the Robotic Arm Dataset (RoAD), a benchmark collected from a robotic arm deployed in a scale-replica smart manufacturing assembly line. The modeling results demonstrate that our method achieves a satisfactory interpretability-simulation accuracy trade-off with up to 0.676 arc-degree simplicity, 0.395 R$^2$ and 0.088 RMSE. In addition, the fault identification results show that the method achieves an F1 score of up to 98.925\%, while maintaining a low conformance checking time of 0.020 seconds, which competes with other deep learning-based methods.

The remainder of the article is structured as follows: \Cref{sec:background} reviews relevant literature on process mining and Petri nets, positioning our contributions and novelty clearly within the existing state-of-the-art. \Cref{sec:methodology} details the proposed process mining-driven method for fault diagnosis. \Cref{sec:evaluation} empirically assesses the methodology's effectiveness using the RoAD benchmark. Finally, \Cref{sec:conclusions} summarizes key insights and outlines potential future research directions.
\section{Background}
\label{sec:background}

Fault diagnosis in \glspl{cps} is intrinsically difficult.  
Hybrid continuous/discrete dynamics, large-scale multivariate sensor streams, and the stringent demand for interpretable decisions in safety-critical domains collectively pose technical and methodological challenges.  
Classical solutions ranging from first-principles physical models and fuzzy rule bases to neural network-driven classifiers have proven effective in specific settings; yet they typically (i) require extensive domain expertise, (ii) offer limited transparency, and (iii) provide little support for realistic behavioral simulation~\cite{interpreting_recurrent_neural_networks_in_process_mining:Hanga:2020}.

To address these limitations, we propose a process mining-driven fault diagnosis method that effectively integrates data-driven learning with the formal structure of Petri nets, whose synergy with data-centric approaches has been demonstrated in prior works~\cite{DASILVEIRA2025103080,LASSOUED2024690,Nadim2022}. 
Unlike black-box models, process mining generates explicit, executable representations of system behavior.  
The remainder of this section introduces the necessary preliminaries on process mining and Petri nets, then surveys existing \gls{cps} fault-diagnosis research to position our contribution.

\subsection{Preliminaries}
\label{sub:preliminaries}

Process mining comprises two core activities: \emph{process discovery} and \emph{conformance checking}~\cite{aalst2022pmhb}.  
Process discovery builds a process model (here, a Petri net) from historical event logs, whereas conformance checking quantifies how closely new logs align with a reference Petri net.  
An event log is a collection of \emph{cases}; each case records a time-ordered sequence of \emph{events} corresponding to state transitions in the underlying \gls{cps}.

\begin{definition}[Event, trace, case, event log]
Let $\mathcal{E}$ be the universe of events, $\mathcal{ST}$ the universe of state transitions, and $\mathcal{T}$ the universe of timestamps.  
Given $e\!\in\!\mathcal{E}$, denote its state transition and timestamp by $\#{\!}_{st}(e)\!\in\!\mathcal{ST}$ and $\#{\!}_{t}(e)\!\in\!\mathcal{T}$, respectively.  
A \emph{trace} is an ordered sequence $\sigma=\langle e_1,\dots,e_k\rangle$ with $e_i\!\in\!\mathcal{E}$; the set of all traces is $\mathcal{E}^{\ast}$.  
Let $\mathcal{C}$ denote the universe of cases and $\#{\!}_{\text{trace}}(c)\!\in\!\mathcal{E}^{\ast}$ the trace associated with case $c\!\in\!\mathcal{C}$.  
An \emph{event log} is a finite subset $L\subseteq\mathcal{C}$.
\end{definition}

In this study, we focus on the control-flow relations and timing distributions extracted from event logs and captured via labeled accepting Petri nets.

\begin{definition}[Labeled accepting Petri net]
\label{def:pn}
Let $P$ and $Tr$ be disjoint node sets ($P\cap Tr=\emptyset$) in a bipartite graph, and let  
$F\subseteq (P\times Tr)\cup(Tr\times P)$ be the set of directed arcs. A Petri net is the triple $(P,Tr,F)$. Its marking $M\in\mathcal{B}(P)$ is a multiset of tokens and encodes the current state. Petri nets that have an initial marking $M_0$ and a final marking $M_{f}$, i.e., an initial and final state, are called \emph{accepting Petri nets}.
Furthermore, consider a set of state transitions $ST\subseteq\mathcal{ST}$ and a labeling function $l_{Tr}:Tr\!\rightarrow\! ST\cup\{\tau\}$, where $\tau$ denotes a silent (unobservable) transition.  
A \emph{labeled accepting Petri net} is the tuple $(P,Tr,F,M_0, M_f,ST,l_{Tr})$.
$\mathcal{N}$ denotes the universe of all labeled accepting Petri nets, hereafter referred to as Petri nets.
\end{definition}

The formal semantics of Petri nets include the ability to \textit{fire} a transition. Given the number of incoming arcs of a transition, if the current marking is such that there is at least a token in each place connected to the transition, it can fire, i.e., the transition consumes one token from each incoming place and generates one token for each outgoing arc. This firing dynamic enables the Petri net to evolve its state, allowing for the above-mentioned conformance checking and simulation. In addition, it is worth noting that there is a special class of Petri nets, namely workflow Petri nets, which have two specific places: a source place and a sink place. In these Petri nets, $M_0$ assigns a token to the source place. Any complete firing sequence moves the Petri net state from $M_0$ to $M_f$, which consists of at least a token in the sink place. Notably, a desirable property of Petri nets is the \emph{soundness}. A Petri net is sound if any firing sequence from the initial marking can reach a unique final marking without leaving dead transitions~\cite{aalst2022pmhb}.

\begin{figure*}[!t]
    \centering

    \begin{minipage}[b]{0.54\textwidth}
        \centering
        \includegraphics[width=\linewidth]{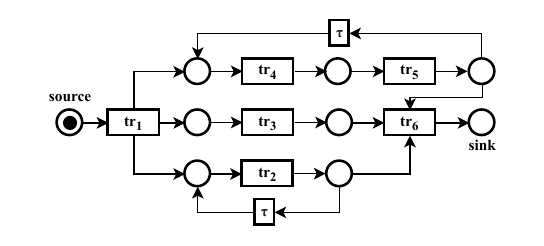}
    \end{minipage}
    \begin{minipage}[b]{0.44\textwidth}
        \centering
        \includegraphics[width=\linewidth]{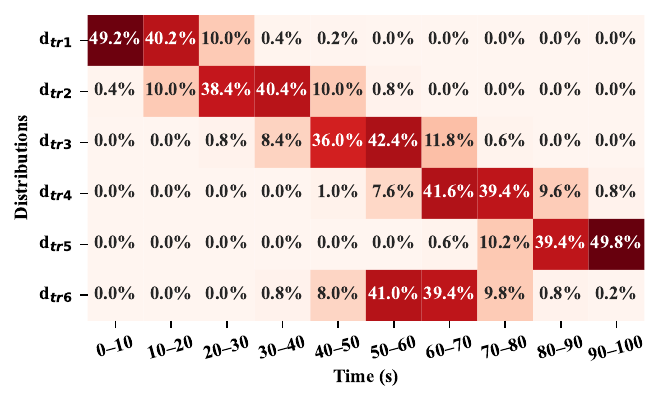}
    \end{minipage}

    \caption{Illustrative workflow stochastic Petri net. The left figure shows the control-flow structure, while the right figure depicts the time perspective.}
    \label{fig:spn}
\end{figure*}
When each transition of a Petri net is further annotated with a firing rate, the model becomes a \emph{stochastic Petri net}~\cite{wittbold2023stochasticdpn}.

\begin{definition}[Stochastic Petri net]
Given a Petri net $N=(P,Tr,F,M_0, M_f,ST,l_{Tr})\!\in\!\mathcal{N}$, let $\mathcal{D}$ be a set of probability distributions and $\delta:Tr\!\rightarrow\!\mathcal{D}$ assign each transition $tr\!\in\!Tr$ its firing-time distribution $d_{tr}$.  
The triple $N_S=(N,\delta,\mathcal{D})$ is termed a \emph{stochastic Petri net}.
\end{definition}

Figure~\ref{fig:spn} shows an illustrative workflow stochastic Petri net. In particular, the left part shows the control-flow view of the Petri net, consisting of transitions $tr_{1},\dots,tr_{6}$, two silent ($\tau$) transitions, the source and sink places, and intermediate places. The control-flow structure constrains the allowable traces of the Petri net, which can be simulated with the firing logic described above. For example, trace $\langle tr_1, tr_2, tr_3, tr_4, tr_5, tr_2, tr_6\rangle$ can be simulated since it is a legal firing sequence. The process is as follows. $tr_1$ fires, producing tokens that enable $tr_2$, $tr_3$, and $tr_4$. $tr_2$ and $tr_3$ partially enable $tr_6$, while $tr_4$ enables $tr_5$, which also partially enables $tr_6$. After $\tau$ fires, $tr_2$ can fire again, further enabling $tr_6$. Finally, $tr_6$ fires, using tokens from $tr_2$, $tr_3$, and $tr_5$, to produce the final token and complete the process. Clearly, when more transitions are concurrently enabled, the simulation engine must resolve race conditions~\cite{roggesolti2014spnsimulation}.

In addition to the control-flow structure, the time perspective, shown on the right part, allows integrating a time duration into the simulations. For example, the firing-time distribution $d_{tr_1}$ is such that the time associated with $tr_1$ firing may last either 0 to 10 seconds with 49.2\% probability, 10 to 20 seconds with 40.2\% probability, 20 to 30 seconds with 10.0\% probability, 30 to 40 seconds with 0.4\% probability, or 40 to 50 seconds with 0.2\% probability. The enhancement of the control-flow structure with the time perspective allows reducing the gap between the Petri net and the actual \gls{cps} dynamics.

\subsection{Related works}
\label{sub:related_works}
To further contextualize our contribution, we revisit traditional model-based techniques, followed by hybrid approaches that attempt to incorporate data-driven insights into Petri net-based fault diagnosis.

\subsubsection{Model-based approaches}
\label{sub:model-based_approaches}
Early research on fault diagnosis in industrial systems proposed Petri nets as a reasoning tool to understand the root causes of system-wide failures~\cite{mansour2013pnfd}.
Subsequently, research aimed to synchronize the fault-free and faulty behaviors within the same Petri net, enabling controllers to identify faults and enforce recovery actions to ensure system safety.
For example, Liu et al.~\cite{liu2013cpshierarchicalfaulterrorfailuredependencies} modeled the fault-error-failure chain of a home \gls{cps} using a stochastic Petri net, pairing model elements with the layered view of the system.
Furthermore, Nazemzadeh et al.~\cite{nazemzadeh2013pndesfaultmodeling} proposed a method to integrate faults in fault-free Petri nets.
They defined fault models as tuples of Petri net elements, including faulty transitions that, when activated, lead to errors.
The diagnosability of faults on Petri nets depends on whether the observed behavior provides sufficient information to identify the activation of faulty transitions. To check if observed behavior causes the activation of faulty transitions, a diagnoser (a model-checking routine) determines the activation of faulty transitions.
Several proposals have introduced different diagnosers based on assumptions about the reference Petri net (e.g., transition types and labeling) and the model-checking routine used (e.g., integer linear programming)~\cite{cabasino2011labeledpndiagnosis,bonafin2022desfd, liu2024popnbnfd}.

\subsubsection{Hybrid approaches}
\label{sub:hybrid_approaches}

Modeling Petri nets for complex \glspl{cps} is challenging and error-prone~\cite{bourdeau2019energyconsumption}. Consequently, researchers have begun to leverage data-driven insights to integrate faulty behavior into fault-free Petri nets. For example, Zhu et al.~\cite{zhu2019desfaultidentification} formulated a set of integer linear programming constraints to incorporate faulty behavior into fault-free Petri nets and identify the root causes of faults.
Similarly, Hou et al.~\cite{hou2021pnfaultidentification} proposed an alternative data-driven method to discover faulty Petri nets.
This method utilizes transition incidence matrices obtained from system traces to address the NP-hard nature of integer linear programming approaches. 

\begin{table*}[!tbp]
  \centering
  \caption{Comparative summary of research on Petri net-based hybrid fault diagnosis, classified according to data type, research area, fault diagnosis method, and simulation support.}
  \vspace{0.5em}
  \resizebox{.95\textwidth}{!}{\begin{tabular}{lllll}
\hline
\textbf{Reference}                              & \textbf{Data type}    & \textbf{Research area}                                                                                 & \textbf{Simulation} & \textbf{Fault diagnosis method}                                                                                                                                                                        \\ \hline
\cite{zhu2019desfaultidentification}            & High-level event data & \begin{tabular}[c]{@{}l@{}}Integer linear \\ programming\end{tabular}                                  & No                  & \begin{tabular}[c]{@{}l@{}}Data-driven extension of a fault-free Petri net\\ to include faulty transitions\end{tabular}                                                                                \\ \hline
\cite{hou2021pnfaultidentification}             & High-level event data & \begin{tabular}[c]{@{}l@{}}Transition incidence\\ matrices\end{tabular}                                & No                  & \begin{tabular}[c]{@{}l@{}}Data-driven extension of a fault-free Petri net\\ to include faulty transitions\end{tabular}                                                                                \\ \hline
\cite{neshastegaran2022planttopologyextraction} & High-level event data & Process discovery                                                                                      & No                  & \begin{tabular}[c]{@{}l@{}}Data-driven modeling of the overall behavior of\\ an industrial system with the Petri net to reason\\ about the root causes of faulty behavior\end{tabular}                 \\ \hline
\cite{friederich2022ddpmreliabilitymodeling}    & High-level event data & Process discovery                                                                                      & No                  & \begin{tabular}[c]{@{}l@{}}Data-driven modeling of the normal behavior of \\ an industrial system with the Petri net and \\ ad-hoc introduction of faulty behavior into\\ the model\end{tabular}       \\ \hline
\cite{wang2022maddc}                            & High-level event data & \begin{tabular}[c]{@{}l@{}}Process discovery\\ Conformance checking\end{tabular}                       & No                  & \begin{tabular}[c]{@{}l@{}}Data-driven modeling of the normal behavior of \\ information systems with the Petri net and \\ verification of deviations from the model of \\ run-time data\end{tabular}  \\ \hline
\cite{vitale2023pmunsupervisedadiiot}           & Low-level sensor data & \begin{tabular}[c]{@{}l@{}}Process discovery\\ Conformance checking\end{tabular}                       & No                  & \begin{tabular}[c]{@{}l@{}}Data-driven modeling of the normal behavior of \\ an industrial system with the Petri net and \\ verification of deviations from the model of \\ run-time data\end{tabular} \\ \hline
\cite{shi2024industrialpmad}                    & Low-level sensor data & Process discovery                                                                                      & No                  & \begin{tabular}[c]{@{}l@{}}Data-driven modeling of the normal behavior of \\ an industrial system with the Petri net and \\ verification of deviations from the model of \\ run-time data\end{tabular} \\ \hline
\cite{debenedictis2023dtadiiot}                 & High-level event data & Conformance checking                                                                                   & Yes                 & \begin{tabular}[c]{@{}l@{}}Verification of deviations from the Petri net of\\ run-time data\end{tabular}                                                                                               \\ \hline
\cite{beyel2024cpsbehavioranalysis}             & High-level event data & \begin{tabular}[c]{@{}l@{}}Process discovery\\ Conformance checking\\ Process enhancement\end{tabular} & No                  & \begin{tabular}[c]{@{}l@{}}Data-driven modeling of the overall behavior of\\ an industrial syustem with the Petri net to reason\\ about the root causes of faulty behavior\end{tabular}                \\ \hline
\cite{vitale2025pmdt}                           & Low-level sensor data & \begin{tabular}[c]{@{}l@{}}Process discovery\\ Conformance checking\end{tabular}                       & Yes                 & \begin{tabular}[c]{@{}l@{}}Data-driven modeling of the normal behavior of \\ an industrial system with the Petri net and \\ verification of deviations from the model of \\ run-time data\end{tabular} \\ \hline
\rowcolor[HTML]{EFEFEF} 
\textbf{Our work}                               & Low-level sensor data & \begin{tabular}[c]{@{}l@{}}Process discovery\\ Conformance checking\\ Process enhancement\end{tabular} & Yes                 & \begin{tabular}[c]{@{}l@{}}Data-driven modeling of the faulty behavior of\\ an industrial system with the Petri net to reason\\ about the root causes of faulty behavior\end{tabular}                  \\ \hline
\end{tabular}}
  \label{tab:hybrid_approaches}
\end{table*}

More recently, process mining has been applied to fault diagnosis and simulation. 
Neshastegaran et al.~\cite{neshastegaran2022planttopologyextraction} developed a topology of an industrial plant through process discovery, leveraging monitored events to enhance the understanding of potential faults.
Friederich et al.~\cite{friederich2022ddpmreliabilitymodeling} proposed a process mining-based framework for reliability modeling of smart manufacturing processes.
The framework integrates the fault-free Petri net obtained through process discovery with custom fault models for each manufacturing resource.
Wang et al.~\cite{wang2022maddc} propose a method that combines neural network-based anomaly detection with process mining. Their method first involves modeling normal behavior using a Petri net from discrete logs collected, for example, from network data and software executions. Next, they perform neural network-based anomaly detection to identify anomalous traces in run-time event logs. Finally, the anomalous traces are checked against the normal behavior to analyze the deviations and provide an explanation of the deviating behavior.
Vitale et al.~\cite{vitale2023pmunsupervisedadiiot} and Shi et al.~\cite{shi2024industrialpmad} proposed using process discovery and conformance checking to characterize normal behavior from low-level sensor data of industrial systems through Petri nets and check deviations from such behavior.
They proposed various unsupervised techniques to discretize sensor data, enabling the use of process discovery and conformance checking.
De Benedictis et al.~\cite{debenedictis2023dtadiiot}
used conformance checking to evaluate whether digital twins could detect new runtime faults.
These faults arise from faulty simulations generated from a Petri net of a proof-of-concept Industrial Internet of Things application.
Beyel et al.~\cite{beyel2024cpsbehavioranalysis} introduced a method of connecting high-level event data from a \gls{cps} in the automotive domain with Petri nets developed through process discovery.
They demonstrated how this method enables the analysis of time spent in \gls{cps} states and the frequency of the transitions between them over time.
Vitale et al.~\cite{vitale2025pmdt} provided an unsupervised approach to characterizing the state transitions of an industrial \gls{cps} involving a real water distribution testbed case study. Their approach involved characterizing the normal \gls{cps} behavior through process discovery and checking any deviations from it through conformance checking.

\subsubsection{Novelty of our work} 
Table~\ref{tab:hybrid_approaches} characterizes the reviewed hybrid approaches according to the data type used, the approach followed, the fault diagnosis method, and whether Petri net-based simulations were performed.

\paragraph{Data type}Many prior works mostly dealt with high-level event data annotated with the specific activity being carried out in the target industrial system. While labeled data allows for the seamless connection of actual events with the system under observation, its availability is limited in real-world applications. Unlike these works, our proposal deals with \emph{low-level sensor data} and extracts states and state transitions unsupervisedly.

\paragraph{Research area}Earlier approaches primarily extended manually designed fault-free Petri nets by integrating faulty transitions with normal behavior. More recent works have shifted toward Petri net–based fault diagnosis through process discovery, conformance checking, or a combination of both. Our work aligns with this emerging direction by integrating discovery, conformance checking, and process enhancement in a unified method.

\paragraph{Simulation}Petri net-based simulation for industrial systems has only been explored by a few of the works we reviewed. Additionally, the simulations performed in these works focus solely on control flow, attempting to replicate the sequences of activities and/or state transitions captured in the Petri net. On the other hand, we aim to integrate the time perspective into our simulations by mining stochastic Petri nets with time distributions for each state transition, thereby allowing for a faithful replication of the actual system's behaviors.

\begin{figure*}[!t]
\centering
\includegraphics[width=0.75\textwidth]{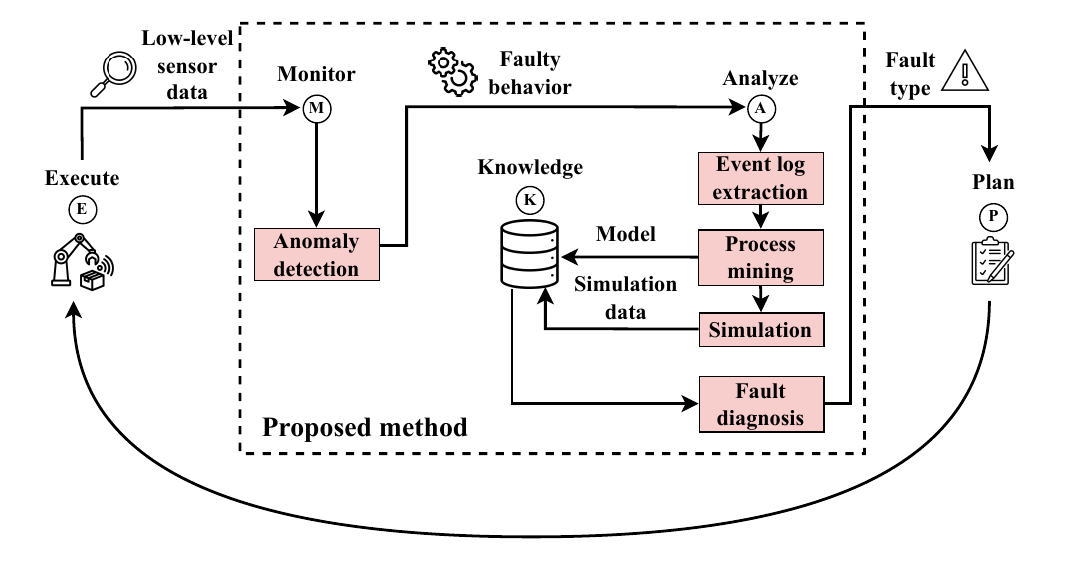}
\caption{The MAPE-K architecture in which our proposed method is contextualized. The red boxes indicate the different phases of our method.}
\label{fig:mape_loop}
\end{figure*}

\paragraph{Fault diagnosis method}Petri net-based fault diagnosis using process mining has largely focused on modeling the normal or overall behavior of a system and then verifying deviations from that model. Such approaches essentially reduce to anomaly detection. In contrast, our method directly captures faulty behaviors and encodes them in specific Petri nets, enabling both fault identification and root-cause analysis. This shift—from detecting deviations to isolating and classifying faults through interpretable Petri nets—distinguishes our approach from earlier work.

\paragraph{Contextualization}We contextualize our proposal within the Monitor-Analyze-Plan-Execute over a shared Knowledge (MAPE-K) architecture (see \Cref{fig:mape_loop}), a well-established framework for autonomous systems~\cite{feng2022mapek}. Our method especially enhances the \textit{Analyze phase} through event log extraction, process mining, simulation, and fault diagnosis, linking offline root-cause analysis with online fault diagnosis. This integration can significantly improve informed decision-making and facilitate proactive corrective actions in manufacturing systems.

\section{Method}
\label{sec:methodology}
\begin{figure*}[!p]
\centering
\includegraphics[width=\textwidth]{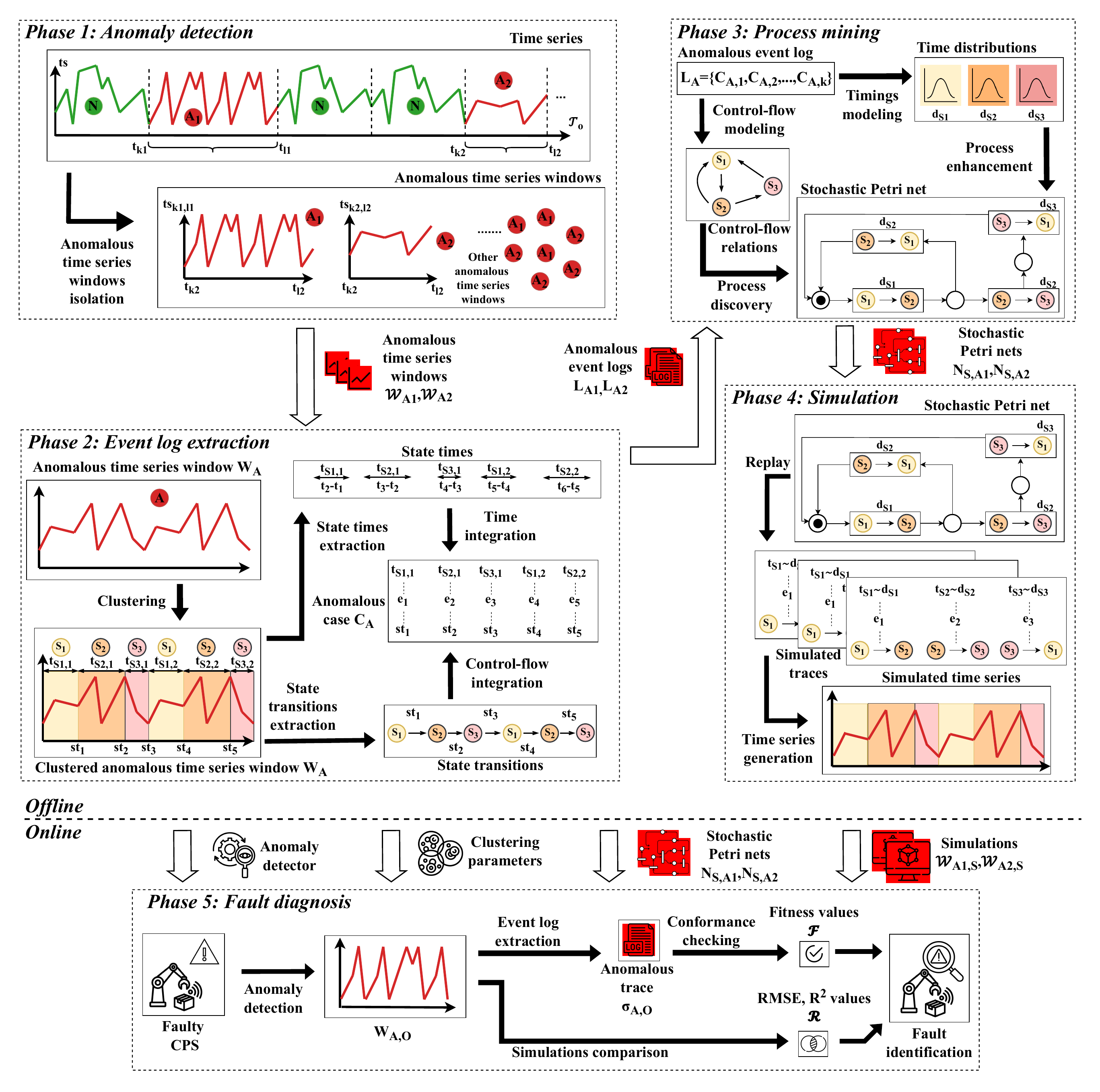}
\caption{
The process mining-driven method for fault diagnosis, organized into an offline part and an online part. The offline part involves four phases, which aim to 1) isolate anomalous time series windows; 2) extract state transitions and state times unsupervisedly; 3) mine stochastic Petri nets characterizing faulty behavior; and 4) simulate the Petri nets to analyze the faulty behavior and support fault identification. Online fault diagnosis uses the outputs of the four phases to isolate and identify \gls{cps} faults.
}
\label{fig:methodology}
\end{figure*}
This section describes the proposed process mining-driven method for fault diagnosis.
The method is split into an offline part and an online part. The offline part is organized into four phases, aiming to capture different fault types into stochastic Petri nets. These are used in the online fault diagnosis phase, which performs fault diagnosis by comparing anomalous behavior with both the structure of the stochastic Petri nets and their offline-generated simulations. The five phases are illustrated in Figure \ref{fig:methodology} and are detailed in the following sections.
\subsection{Phase 1: Anomaly detection}
The objective of this phase is to isolate anomalous behavior from normal behavior in manufacturing \glspl{cps}. Notably, these systems collect large amounts of low-level sensor data in the form of multivariate time series, which can include anomalies~\cite{farahani2023tspatternrecognition}. 
As a result, this phase specifically involves detecting anomalous behavior from these data.

\begin{definition}[Multivariate time series]
Let us denote $\mathcal{T}=\{t_i:i\in\{0,\dots,o-1\},o\in\mathbb{N}\}$ a set of timesteps up to $o\in \mathbb{N}$. A multivariate time series $ts$ with $p$ features is defined as follows.
\begin{flalign}
&ts=\{(t_i,ts(t_i)):t_i\in\mathcal{T}, x(t_i)\in\mathbb{R}^p\}.&
\end{flalign}
In the following, we refer to a multivariate time series of $o$ samples and $p$ features as $ts\in\mathbb{R}^{o\times p}$
\end{definition}

Multivariate time series anomaly detection has become highly effective, especially due to advancements in deep learning approaches that can accurately discriminate normal patterns from anomalous ones~\cite{elia2025adsmartmanufacturing, darban2025dltsad, chatterjee2022adsurvey, cook2020iottsadsurvey}.
While these approaches have achieved excellent results, they are unable to characterize and classify faults, as their objective is to model normal patterns and flag anomalous observations. 
Phase 1 builds on these abilities and isolates collective anomalies in time series, defined as sequences of contiguous samples that deviate from expected patterns~\cite{chandola2012addiscretesequencessurvey, qin2023collectivead}.
We refer to such sequences as \textbf{anomalous time series windows}. 
    
\begin{definition}[Anomalous time series window]
Let $ts\in \mathbb{R}^{o\times p}$ be a multivariate time series with $p$ features and $o$ samples. Let $ts(i)\in\mathbb{R}^p$ be a time series sample at timestep $t_i\in\mathcal{T}$. Let $t_k,t_l\in\mathcal{T},k<l\leq o$. A \emph{time series window} $ts_{k,l}$ is the set:
\begin{flalign}
&ts_{k,l}=\{ts(t_j)\in\mathbb{R}^p:t_j\in\mathcal{T},t_k\leq t_j \leq t_{l}\}\subseteq ts.&
\end{flalign}
In the following, we refer to an anomalous time series window of anomaly $A$ as $W_A\subseteq ts$ and denote $\mathcal{W}_{A}=\{W_{A,1},W_{A,2},\dots,W_{A,\psi}\}$ the set of $\psi$ anomalous time series windows of anomaly $A$ isolated from $ts$.
\end{definition}

The sets $\mathcal{W}_{A_1},\mathcal{W}_{A_2},\dots,\mathcal{W}_{A_m}$ capture $m$ different fault types. Figure \ref{fig:methodology} demonstrates this concept by decomposing a historical time series $ts$ into multiple time series windows, where a normal pattern is disrupted by anomalies $A_1$ and $A_2$. These are isolated and collected into sets $\mathcal{W}_{A_1}$ and $\mathcal{W}_{A_2}$, which are passed on to the next step for event log extraction.
    
\subsection{Phase 2: Event log extraction}
The objective of this phase is the processing of low-level sensor data and their conversion into event logs so that process mining algorithms can be applied. To this aim, the phase first involves characterizing data as sequences of events. Common approaches for discretizing time series data and extracting event logs involve grouping each sample into clusters \cite{hemmer2021adpm, vitale2023pmunsupervisedadiiot, su2025pmsd} or by identifying specific events, such as increasing or decreasing trends and specific activities \cite{vaneck2016pmsd, shi2024industrialpmad, vitale2025pmdt}. While the latter approaches are more precise in identifying events from sensor data, they require domain knowledge and, in some cases, accurate labeling. On the other hand, clustering-based approaches are unsupervised and very valuable in contexts where the collected data is not labeled or the manual identification of possible trends is impractical. Hence, our proposal adopts the first class of methods and clusters time series values into a set of \textbf{states} and collects events by identifying transitions between clusters, i.e., the \textbf{state transitions}.
    
\begin{definition}[State and state transition]
\label{def:state_transition}
Let $W_A\subseteq ts\in\mathbb{R}^{o\times p}$ be an anomalous time series window isolated from a and $\mathcal{S}=\{s_1,\dots,s_\alpha: s_i\in\mathbb{R}^p,\alpha\in\mathbb{N}\}$ be a set of clusters obtained by assigning $W_A$ values to $\alpha$ centroids. We define $s_t\in \mathcal{S},t\in\mathcal{T}_o$ the \emph{state} associated with $ts(t)\in W_A$. Therefore, $\mathcal{S}$ is the set of states. Given $t_k,t_{l}:0\leq k<l\leq o$, there is a \emph{state transition} associated with $t_{l}$ if $\forall t_j:t_k<t_j<t_{l},s_{t_j}=s_{t_k}$ and $s_{t_k}\neq s_{t_l}$. An event $e$ associated with the pair of states $s_{t_k},s_{t_l}$ is such that $\#_{st}(e)=st_l=s_{t_k}\rightarrow s_{t_l}$ and $\#_{t}(e)=t_l$. The state $s_{t_k}$ is associated with $st_l$.
\end{definition}

The application of this definition requires extracting the set of states $\mathcal{S}$ by applying a suitable clustering algorithm to the original time series $ts$ from which the anomalous time series windows are isolated. Numerous such algorithms are documented in the literature. Arguably, an increasing number of clusters leads to a more complex state space with a possibly better discretization of the time series.

Figure \ref{fig:methodology} shows how a given anomalous time series window $W_A$ gets discretized through clustering into an alternation of three different states, namely $s_1$, $s_2$, and $s_3$. Each state $s$ has a given duration $t_{s}$ that depends on the time instant at which a state transition occurs. To integrate the time perspective into the Petri nets, the \textbf{state times} are collected.

\begin{definition}[State time]
Let $\mathcal{S}$ be a set of states, $e\in\mathcal{E}$ be an event, and $\#_{st}(e)=s_{t_k}\rightarrow s_{t_l}$ a state transition. The \emph{state time} of $s_{t_k}\in\mathcal{S}$ associated with $\#_{st}(e)$ is $t_l-t_k$.
\end{definition}

Each anomalous time series window $W_A$ corresponds to an anomalous case $C_A$. $C_A$ collects the sequence of state transitions in $W_A$ and the associated state times. The set of cases collected from set $\mathcal{W}_A$ composes the anomalous event log $L_A$. In conclusion, the anomalous event logs $L_{A_1}$ and $L_{A_2}$ of anomalies $A_1$ and $A_2$ are passed on to phase 3.

\subsection{Phase 3: Process mining}
The objective of this phase is to encode the anomalous event logs into stochastic Petri nets to build a fault dictionary. To this aim, the anomalous event log $L_A$ of a given anomaly $A$ is processed in two different ways through process discovery and process enhancement.

\subsubsection{Process discovery}
To discover the control-flow structure, the relations between the different state transitions within the cases of $L_A$ are captured. For example, by analyzing all the cases of $L_A$, it may result that states $s_1$ and $s_2$ repeat in a loop, state $s_3$ always follows $s_2$, and $s_1$ always follows $s_3$. Process discovery can encode these control-flow relations into a Petri net, which ultimately represents the \gls{cps} behavior under faulty conditions.

\begin{definition}[Process discovery algorithm]
Let $L\subseteq\mathcal{C}$ be an anomalous event log. A \emph{process discovery algorithm} is a function $\gamma$ that captures the control-flow relations among the state transitions of the cases of $L$ and encodes them to a Petri net $N\in\mathcal{N}$: $\gamma(L)=N$.
\end{definition}

The set of control-flow relations discovered by $\gamma$ depends on the specific process discovery algorithm being used.
For example, the inductive miner~\cite{leemans2013im} identifies several types of control-flow relations, such as sequence, parallel, exclusive, and redo-loop relations.
It does this by splitting the event log into simpler segments, which enables the recognition of these relations.
The integer linear programming-based miner \cite{vanzelst2018ilp} finds causal relations between the activities of an event log and specifies a system of inequalities to find the so-called regions, i.e., two sets of activities such that the prior execution of the first set enables the execution of the second set. 
The heuristics miner \cite{weijters2011hm} finds dependency relations between the activities of an event log through the dependency graph, which collects frequency-based causal relations between activities based on their occurrences in the event log.

Notably, the inductive miner, integer linear programming-based miner, and heuristics miner discover workflow Petri nets. However, the types of control-flow relations identified by the process discovery algorithm determine the range of possible Petri nets the algorithm can generate.
This constraint is known as the \emph{representational bias}, and can significantly impact the quality of the resulting Petri net~\cite{aalst2022pmhb}. In turn, this could impact the ability to adequately characterize the faulty behavior.

The aforementioned biases may cause two different phenomena in the resulting Petri nets: overfitting and underfitting. For example, although the inductive miner builds perfectly fitting Petri nets from the data, it forces this property by introducing many silent transitions. This results in the Petri net being underfit, i.e., it introduces many more behaviors than those present in the observed data. Yet, while the integer linear programming-based miner limits the introduction of silent transitions, it enforces many constraints in building the above-mentioned regions through the system of inequalities, thereby leading to overfitting Petri nets; i.e., the model is very specific to the data at hand. Finally, the heuristics miner may also lead to overfitting Petri nets by attempting to capture all the causal relations, including the infrequent ones. To mitigate the underfitting and overfitting phenomena, the algorithms were extended with filtering mechanisms that allow reducing the impact on the resulting Petri net~\cite{aalst2022pmhb}.

In our study, we are particularly interested in reducing the overfitting effect while maintaining a good degree of specificity, i.e., avoiding underfitting. By attempting to model the behavior of complex and noisy event logs, the resulting Petri net may have numerous places and arcs, resulting in an overwhelming structure whose interpretation can be cumbersome and challenging. To quantitatively evaluate the interpretability of the Petri net, the arc-degree simplicity can be used~\cite{vazquez2015prodigen}.

\begin{definition}[Arc-degree simplicity]
Let $N\in\mathcal{N}$ be a Petri net and $arc_{P,N}, arc_{Tr,N}\in\mathbb{N}$ the total number of incoming and outgoing arcs in $N$ related to, respectively, the set $P$ of places and the set $Tr$ of transitions. Let $arc_N\in arc_{P,N}\cup arc_{Tr,N}$ be the arc degree of $N$. The arc-degree simplicity $S_{arc}$ of $N$ is:
\begin{flalign}
\label{eq:s_arc}
    &S_{arc}(N)= \frac{1}{1+(\overline{arc}-2)}, \overline{arc}=\frac{1}{|arc_N|}\sum_{x\in arc_N}x.&
\end{flalign}
\end{definition}

According to the arc-degree simplicity, the simplest Petri net ($S_{arc}=1$) has exactly one incoming and one outgoing arc per node. The more complex the dynamics, the more incoming and outgoing arcs per node there are, the closer to 0 $S_{arc}$ becomes, and the less interpretable the Petri net is.
    
\subsubsection{Process enhancement}
Once a quality Petri net is built with process discovery, process enhancement is performed to integrate the time perspective into the Petri net~\cite{aalst2022pmhb}. Such enhancement involves collecting the \textbf{state time distributions} from the cases of the anomalous event log.

\begin{definition}[State time distribution]
\label{def:state_distribution}
Let $L_A$ be the anomalous event log associated with anomaly $A$ and $s\in\mathcal{S}$ a state. Let $T_s=\{t_{s,e}\in\mathbb{R}:\#_{st}(e)=s\rightarrow \bar{s}, e\in L_A\}$ be the set of state times associated with $s$ found in the events of the cases of $L_A$. The state time distribution $d_s:\mathbb{R}\rightarrow\mathbb{R}$ is a function that models the probability based on $T_s$ that $s$ lasts for a certain time.
\end{definition}

The integration of the state time distributions in the Petri net, as illustrated in Section \ref{sec:background}, allows adding the time perspective. In the case of Figure \ref{fig:methodology}, the three states have time distributions $d_{s_1}$, $d_{s_2}$, and $d_{s_3}$ assigned to each of the four transitions. The resulting enhanced model constitutes a \textbf{stochastic Petri net} that supports simulation and quantitative analysis of the temporal behavior of the process. By combining process discovery with temporal enhancement, the model provides a more realistic and analytically powerful representation of the faulty \gls{cps}, enabling both diagnosis and predictive evaluation of anomalous system dynamics. In conclusion, the stochastic Petri nets $N_{S,A_1}$ and $N_{S,A_2}$ of anomalies $A_1$ and $A_2$ are passed on to phase 4.
    
\subsection{Phase 4: Simulation}
The objective of this phase involves simulating the Petri net to analyze the faulty behavior and provide synthetic yet realistic anomalous time series windows.

First, the simulation \textbf{replays} a stochastic Petri net to generate new traces.
A simulated trace is of the form $\langle st_1,\dots,st_f\rangle$.
A state is associated with each state transition of the trace. 
E.g, if $st_1=s \rightarrow \bar{s}$, with $s_\beta$ and $s_\gamma$ being two states of the \gls{cps}, then $s$ is associated with $st_1$.
At each state in the simulated trace, a corresponding state time is randomly selected from the state time distribution linked with that transition.
As a result, the sequence of state transitions is transformed into state-state time pairs.
The sequence of state-state time pairs is the \textbf{simulated time series window}.
    
\begin{definition}[Simulated time series window]
\label{def:sim_ts}
Let $N\in\mathcal{N}$ be a Petri net, $\mathcal{S}=\{s_1,\dots,s_\alpha: s_i\in\mathbb{R}^p,\alpha\in\mathbb{N}\}$ a set of $\alpha$ states, $\mathcal{D}=\{d_1,\dots, d_\alpha: d_i:\mathbb{R}\rightarrow\mathbb{R},\alpha\in\mathbb{N}\}$ the set of state time distributions, $\delta$ the function that associates the state time distributions with the state transitions of $N$, and $N_S=(N,\delta,\mathcal{D})$ a stochastic Petri net. Let $\sigma_{sim}=\langle st_1,\dots, st_k \rangle$ be a trace of $k$ state transitions obtained by replaying $N_S$. The corresponding \emph{simulated time series window} is the sequence of pairs $\langle (s_{st_1},t_{st_1}),\dots, (s_{st_k}, t_{st_k})\rangle$. The $i$-th pair $(s_{st_i}, t_{st_i})$ captures the state $s_{st_i}\in \mathcal{S}$ associated with $st_i$ and the corresponding state time $t_{st_i}$ sampled from $\mathcal{D}$.
\end{definition}

It is worth mentioning that, as explained in Section \ref{sec:background}, the generated traces --- and their conversion into time series windows using the approach above --- depend on both the control-flow structure and the time perspective. While the control-flow structure may fit the faulty behavior, the set of allowable traces can be broader than those actually seen in the data, as the process discovery algorithms attempt to capture as much structure as possible. Likewise, the state time distributions are random variables that probabilistically sample a time to associate with states. Hence, the resulting traces capture the stochastic essence of the anomalous process, striking a balance between model generality and realistic temporal variability.

\subsection{Phase 5: Fault diagnosis}
The online part of the method involves performing fault diagnosis using the outputs provided by the offline phases, namely the anomaly detector trained during phase 1, the clustering parameters of phase 2, the stochastic Petri nets of phase 3, and the simulations of phase 4.

The faulty \gls{cps} is monitored and the anomaly detector used to isolate the anomalous time series window $W_{A,O}$ from the normal behavior. Next, after the anomalous trace $\sigma_{A,O}$ is obtained by applying event log extraction using the same clustering parameters as those obtained during phase 2, the fault is diagnosed by combining two different approaches: \textbf{conformance checking} and \textbf{simulations comparison}.

Conformance checking involves evaluating the degree of similarity between a trace and the structure of the stochastic Petri net. There are various algorithms for performing such checks. The state-of-the-art conformance checking algorithm is the alignment-based one \cite{aalst2022pmhb}. This approach attempts to find the best path across the reference stochastic Petri net that most accurately approximates the traces in the event log that deviate from normal control flow. This is achieved by considering the log moves in an event log and their alignment with the reference stochastic Petri net, as determined by evaluating the corresponding model moves. This alignment is compared to the worst-case alignment, and the fitness is calculated using a cost function $\eta$ that quantifies the alignment's penalty for each misaligned pair of log and model moves.

\begin{figure*}[!t]
    \centering
    \includegraphics[width=\textwidth]{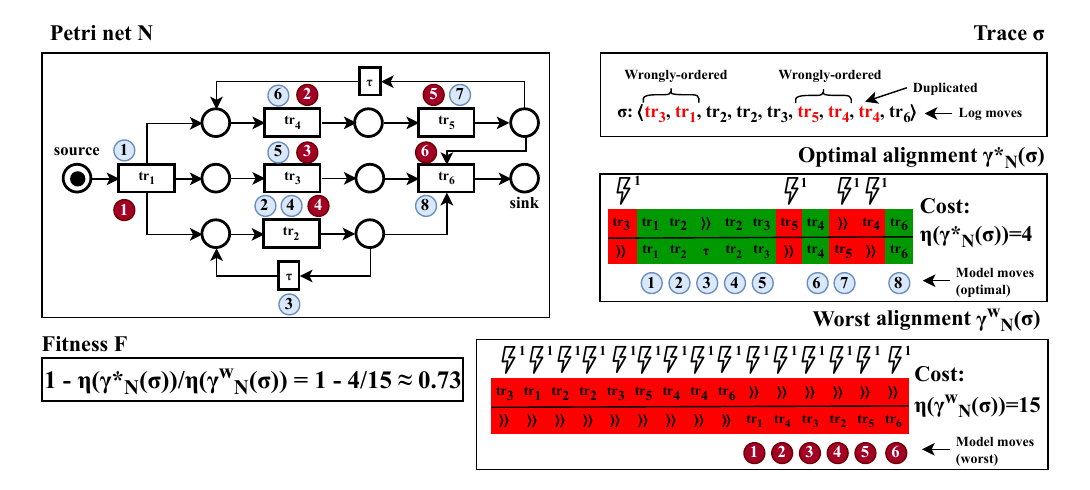}
    \caption{Example fitness calculation between a trace $\sigma$ and the control-flow structure of Petri net N of Figure \ref{fig:spn}.}
    \label{fig:alignment}
\end{figure*}

\begin{definition}[Fitness]
\label{def:fitness}
Let $\sigma_{A,O}$ be the anomalous trace and $N$ a Petri net. Moreover, let $\gamma^*_N(\sigma_{A,O})$ and $\gamma^w_N(\sigma_{A,O})$ be the optimal and worst alignments of $\sigma_{A,O}\in L$ with respect to $N$, respectively. Finally, let $\eta(\gamma^*_N(\sigma_{A,O}))$ and $\eta(\gamma^w_N(\sigma_{A,O}))$ be the costs assigned to the optimal and worst alignments, respectively. The fitness $F_{\sigma_{A,O}}\in[0,1]$ is
\begin{flalign}
\label{eq:fitness}
&F(\sigma{_{A,O}},N)=1-\frac{\eta(\gamma^*_N(\sigma_{A,O}))}{\eta(\gamma^w_N(\sigma_{A,O}))}.&
\end{flalign}
\end{definition}

Figure \ref{fig:alignment} shows an example fitness calculation based on the optimal and worst alignment of an example trace $\sigma=\langle tr_3,tr_1,tr_2,tr_2,tr_3,tr_5,tr_4,tr_4,tr_6\rangle$ with respect to the control-flow structure of the Petri net N shown in Figure \ref{fig:spn}. The trace has several control-flow anomalies, including wrongly-ordered and duplicated transitions. When finding the optimal alignment $\gamma^*_{N}(\sigma)$, these anomalies translate into ``unsynchronized'' moves, i.e., there is no possible matching between the log and model moves of the alignment. By considering the worst alignment as the shortest firing sequence on the model following all the log moves, the costs of the best and worst alignments sum to, respectively, 4 and 15, leading to a fitness $F(\sigma,N)$ of approximately 0.73.

Simulations comparison involves evaluating the difference between the anomalous time series window and the simulations performed offline. Several metrics can be used for this purpose. In our method, we consider two metrics: the Root Mean Squared Error (RMSE) and the coefficient of determination ($R^2$):
\begin{flalign}
\label{eq:rmse}
&\text{RMSE} = \sqrt{\frac{1}{n} \sum_{i=1}^{n} (W_{A,O}(i) - W_{A,S}(i))^2},&
\end{flalign}
\begin{flalign}
\label{eq:r2}
&\text{R}^2 = 1 - \frac{\sum_{i=1}^{n} (W_{A,O}(i) - W_{A,S}(i))^2}{\sum_{i=1}^{n} (W_{A,O}(i) - \overline{W_{A,O}})^2},&
\end{flalign}
where $W_{A,S}$ is a simulated window and $n$ is the number of samples of the windows. While RMSE evaluates the similarity of the simulation to the amplitude of the actual signal, R$^2$ provides information about the difference in explained variance between the actual signal and the simulated window.

The fault identification part of fault diagnosis uses these three metrics according to Algorithm \ref{fi_algo}. The algorithm identifies the best candidates for each of the three metrics and makes a final decision based on the majority among them.

\begin{algorithm}
\caption{Fault identification}
\label{fi_algo}
\begin{algorithmic}[1]
\State \textbf{Input:} $N_{S,A_1},\dots,N_{S,A_m}$, $W_{A_1,S},\dots,W_{A_m,S}$, $W_{A,O}$, $\sigma_{A,O}$
\State \textbf{Output:} Fault type

\State $\mathcal{F}, \mathcal{R}_{\text{RMSE}}, \mathcal{R}_{\text{R}^2} \gets \{\}, \{\}, \{\}$
\For{$i = 1$ to $m$}
    \State $\mathcal{F}, \mathcal{R}_{\text{RMSE}}, \mathcal{R}_{\text{R}^2} \gets 
           \mathcal{F} \cup \{\mathrm{F}(\sigma_{A,O}, N_{S,A_i})\}, \;
           \mathcal{R}_{\text{RMSE}} \cup \{\mathrm{RMSE}(W_{A,O}, W_{A_i,S})\}, \;
           \mathcal{R}_{\text{R}^2} \cup \{\text{R}^2(W_{A,O}, W_{A_i,S})\}$
\EndFor
\State Indices $\gets (\arg\max \mathcal{F}, \arg\min \mathcal{R}_{\text{RMSE}}, \arg\max \mathcal{R}_{\text{R}^2})$
\State Fault type $\gets$ majority(Indices)
\State \Return Fault type
\end{algorithmic}
\end{algorithm}

\newpage{}
\section{Case study}
\label{sec:evaluation}
\begin{figure*}[!t]
    \centering
    \includegraphics[width=\textwidth]{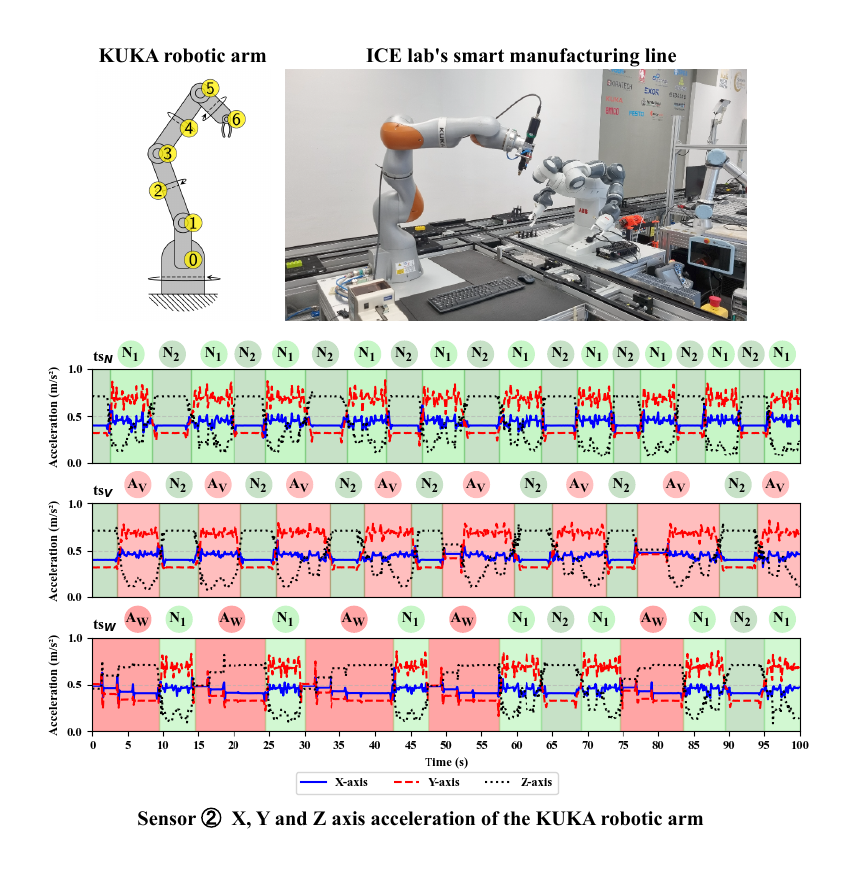}
    \caption{Experimental setup of the Robotic Arm Dataset (RoAD) testbed at the ICE laboratory of the University of Verona. The smart manufacturing line integrates a KUKA robotic arm instrumented with seven sensors at each mechanical joint, measuring the X, Y, and Z-axis acceleration. The sensor 2 acceleration values are captured in three different conditions: normal (ts$_N$), weight anomaly (ts$_W$), and velocity anomaly (ts$_V$). Each time series either show the normal patterns N$_1$ and N$_2$ or anomalous patterns A$_V$ and A$_W$.}
    \label{fig:ice_lab}
\end{figure*}

We apply the process mining-driven fault diagnosis method to the \gls{road} case study~\cite{mascolini2023road}, whose dataset is one of the best representatives in the manufacturing \gls{cps} context, as reported in the work of Golendukhina et al.~\cite{Golendukhina2024manufacturingdatasetreview} and Khan et al.~\cite{dataset_survey:Uzair:2025}. The goals of the experiments involve analyzing:
\begin{itemize}
    \item{The impact of key methodology factors on the unsupervised modeling and simulation quality of faulty \gls{cps} behaviors (modeling and simulation quality).}
    \item{The ability to identify different faulty \gls{cps} behaviors and clearly distinguish them from one another to build a fault dictionary (fault identification).}
\end{itemize}

We evaluate the effectiveness of the methodology in modeling and simulating faulty time series windows collected from the \gls{road} dataset. The discussion below describes the \gls{road} case study and the application of the method to it, providing an overview of the experiments and a detailed discussion of the results, which outline the findings and potential threats to the validity of the experimentation.

\subsection{RoAD dataset}

\subsubsection{Physical Plant} The \gls{road} case study comprises real multivariate time series data collected from various sensors mounted on a Kuka robotic arm, a collaborative robotic arm deployed in a scaled replica of a manufacturing process, as shown in the top-right corner of \Cref{fig:ice_lab}. The robotic arm is integrated into a fully-fledged production line located in the \gls{ice} Laboratory\footnote{\url{https://www.icelab.di.univr.it/laboratory}}, a cutting-edge research facility at the University of Verona dedicated to cyber-physical production systems.
This production line includes a diverse set of industrial-grade machines, representative of modern smart manufacturing environments: a quality control cell with high-resolution laser scanners and cameras for visual inspection, a robotic assembly cell with two collaborative robotic manipulators for flexible handling tasks, a subtractive manufacturing cell equipped with a four-axis computer numerical control milling machine, and a functional control cell that integrates a flying probe system for automated functional testing.
Two automated guided vehicles manage the internal logistics, transporting goods between the warehouse and the conveyor belt, enabling the emulation of complex material flows.
The KUKA robotic arm plays a pivotal role within this system, performing critical operations such as assembly, disassembly, and tightening.
The production line is based on the flexible manufacturing paradigm. 
Rather than adhering to rigid, predefined workflows, it enables the dynamic composition of manufacturing processes by orchestrating a set of reusable processing capabilities offered by each work cell.
This configuration not only mirrors the adaptability required in Industry 4.0 scenarios but also provides a high-fidelity experimental environment for evaluating advanced automation, orchestration strategies, and verification techniques under realistic conditions.

\subsubsection{Sensor data} At the top-left corner of \Cref{fig:ice_lab} is a schematic of the physical plant and the set of sensors, labeled 0 to 6, deployed across its structure. The multivariate time series of each sensor includes measurements from accelerometers, gyroscopes, and power consumption sensors. The recorded multivariate time series reflects both nominal and anomalous behaviors of the KUKA robotic arm. The identified anomalies are categorized into distinct classes. In this study, we focus on the X, Y, and Z axis accelerations recorded by sensor 2, as they clearly capture both regular and anomalous behaviors. At the bottom of \Cref{fig:ice_lab}, three acceleration time series sampled at 10 Hz are shown: ts$_N$, ts$_V$, and ts$_W$. The normal time series, ts$_N$, alternates between two recurring patterns, N$_1$ and N$_2$, at regular intervals of approximately 50 samples (5 seconds). These two patterns correspond to distinct activities within the manufacturing workflow: picking up an object and moving it to another location.
To simulate anomalies, two modified time series were collected: ts$_V$ and ts$_W$, corresponding to velocity and weight anomalies, respectively. The velocity anomaly reduces the robot’s movement speed, affecting pattern N$_1$. This alteration, annotated as A$_V$, manifests as slower dynamics and lower acceleration amplitudes. Conversely, the weight anomaly increases the object’s mass, affecting pattern N$_2$. This alteration, annotated as A$_W$, is characterized by slower motion and a distinctive step-like acceleration pattern.

\paragraph{Method application} We aim to capture the faulty \gls{cps} behavior through our methodology as follows. First, we set out a controlled environment for the experiments by manually tuning the anomaly detection step to fixed accuracy configurations. This allows us to evaluate the impact of the anomaly detection part on the subsequent fault diagnosis phase. 
Next, we apply the event log extraction and process mining steps to either of the weight and velocity time series windows to obtain stochastic Petri nets capturing the faulty \gls{cps} behavior. 
Furthermore, we simulate the reference Petri net a given number of times and evaluate the alignment of the simulated time series windows with the original faulty \gls{cps} behavior. 
Finally, we evaluate the effectiveness of our proposal in identifying different types of faults.

\subsubsection{Experimental setup}
\paragraph{Hardware and Software Specifications} The software for the experiments was implemented using Python and run on a Windows 10 machine with an Intel Core i9-11900K CPU at 3.50 GHz and 32GB of RAM. The software uses machine learning and PM libraries, such as \texttt{scikit-learn} and \texttt{pm4py}. In addition, it includes a set of DOS batch scripts to automatically replicate the experiments carried out in this work, publicly available on GitHub\footnote{\url{https://github.com/francescovitale/pm_based_modeling_simulation}.}. 

\begin{table}[!t]
\centering
\caption{The RoAD dataset properties.}
\label{tab:dataset_props}
\resizebox{0.8\columnwidth}{!}{%
\begin{tabular}{llll}
\hline
\textbf{Property}                     & \textbf{ts$_{N}$} & \textbf{ts$_{V}$} & \textbf{ts$_{W}$} \\ \hline
Samples                               & \multicolumn{3}{l}{10000 ($\approx$ 16 min.)}             \\
Features                              & \multicolumn{3}{l}{3 (X, Y and Z acc.)}                   \\
Windows                               & 161               & 159               & 76                \\
Window size                           & 49                & 62                & 130               \\
Normal windows                        & 161               & 79                & 37                \\
Anomalous windows $\mathcal{W}_{A}$   & 0                 & 80                & 48                \\
Training windows $\mathcal{W}_{A,tr}$ & 0                 & 48                & 23                \\
Test windows $\mathcal{W}_{A,tst}$    & 0                 & 32                & 16                \\ \hline
\end{tabular}%
}
\end{table}

\paragraph{Data pre-processing and simulation}
The sensor data was pre-processed so that controlled experimentation could be carried out. The resulting dataset properties after pre-processing are shown in Table \ref{tab:dataset_props}.
First, ts$_N$, ts$_V$ and ts$_W$ were normalized between 0 and 1. 
Second, 10000 samples for each time series of sensor 2 of the robotic arm were retrieved.
Third, we manually extracted time series windows to obtain a labeled dataset with which we evaluate the method's performance. Specifically: ts$_N$ has a total of 161 normal windows, with an average window size of 49 samples; ts$_V$ has a total of 159 windows, of which 79 are normal and 80 are anomalous, with an average window size of 62 samples; ts$_W$ has a total of 130 windows, of which 37 are normal and 48 are anomalous, with an average window size of 130 samples. We split each set of anomalous windows $\mathcal{W}_A$ into a training set $\mathcal{W}_{A,tr}$ and a test set $\mathcal{W}_{A,tst}$, using a simple 75\% holdout procedure.
Finally, during the simulation, we set the number of traces to simulate to 300 and, for each trace, generate the simulated time series according to Definition \ref{def:sim_ts}. The number of simulated time series windows was sufficient to extract a statistically reliable estimation of how different the simulated Petri net was compared to the actual time series windows.
We repeat this procedure three times to obtain mean and standard deviation statistics for the metrics used to evaluate the modeling and simulation quality and fault identification capabilities.

\subsubsection{Factors} The key factors influencing the method's capabilities involve the anomaly detection accuracy, the clustering algorithm, and the process discovery algorithm. The accuracy of anomaly detection depends on the number of true positives (TP), false positives (FP), true negatives (TN), and false negatives (FN). Positives are anomalous time series windows, whereas negatives are normal time series windows. Therefore, anomaly detection accuracy:
\begin{flalign}
\label{eq:acc}
    &\text{Acc}=\frac{\text{TP}+\text{TN}}{\text{TP}+\text{TN}+\text{FP}+\text{FN}}&
\end{flalign}
is changed by varying the content of the training windows $\mathcal{W}_{A,tr}$. For example, if the accuracy is perfect, the set only contains true positives. If the accuracy is 75\%, 25\% of true positives are replaced with the same percentage of false positives.
The clustering algorithm employed in the experiments to extract the state, state transitions, and state time from the time series is K-means, a well-known clustering technique that associates each data point with a single cluster.
This technique is desirable as it allows the user to specify the desired number of clusters to be extracted. The K factor relates to the number of states and state transitions that can be captured. For example, if K equals 3, the \gls{cps} dynamics are discretized into 3 states and 6 state transitions. Hence, the higher K, the finer-grained the discretization is.
The process discovery techniques used to extract the Petri net from the anomalous event log are the \gls{imf} \cite{leemans2014imf}, \gls{ilp} \cite{vanzelst2018ilp}, and Heuristics Miner (HM) \cite{weijters2011hm}.
These three techniques can tolerate noise in event logs and utilize three distinct approaches to discover control-flow relations, which allows evaluating the representational bias impact outlined in Section \ref{sec:methodology}. To minimize the impact of noise in event logs while minimizing excessive filtering, we set the noise threshold for these three algorithms to 75\%.

\subsubsection{Metrics} The metrics used in the first experiment to evaluate the Petri net quality are the arc-degree simplicity (Equation \ref{eq:s_arc}), RMSE (Equation \ref{eq:rmse}), and R$^2$ (Equation \ref{eq:r2}). The second experiment to evaluate the fault identification capabilities involves the F1-score (F1), calculated as follows. Let $\mathcal{W}_{A_{target}}$ be the anomalous time series windows of the target fault to identify. These are the positive samples. Furthermore, let $\mathcal{W}_{A_{other}}$ be the anomalous time series windows of the other fault. These are the negative samples. F1 is:
\begin{flalign}
&\text{F1}=\frac{2\text{TP}}{2\text{TP}+\text{FP}+\text{FN}}.&    
\end{flalign}
Furthermore, to assess the online capabilities of our method, we include the conformance checking time, which is the time needed to perform alignment-based conformance checking for a single trace.

\begin{table*}[!t]
    \centering
    \caption{The modeling and simulation quality results of the offline part of our fault diagnosis method for both the velocity and weight anomaly. Bold numbers highlight the best R$^2$ per process discovery algorithm and anomaly type. The grey cells highlight the best Petri nets for the velocity and weight anomalies in terms of simplicity ($S_{arc}$), coefficient of determination ($R^2$), and \gls{rmse} trade-off.}
    \vspace{0.5em}
    \resizebox{\textwidth}{!}
    {\begin{tabular}{cllllllllllll}
\hline
\multicolumn{1}{l}{}                                         & \textbf{}  & \multicolumn{3}{l}{\textbf{ILP (75\% noise tolerance)}}                        & \textbf{} & \multicolumn{3}{l}{\textbf{IMf (75\% noise tolerance)}}                                                                                                &  & \multicolumn{3}{l}{\textbf{HM (75\% noise tolerance)}}                         \\ \cline{3-5} \cline{7-9} \cline{11-13} 
\multicolumn{1}{l}{\multirow{-2}{*}{\textbf{}}}              & \textbf{K} & \textbf{S$_{arc}$}       & \textbf{$R^2$}           & \textbf{RMSE}            &           & \textbf{S$_{arc}$}                               & \textbf{$R^2$}                                   & \textbf{RMSE}                                    &  & \textbf{S$_{arc}$}       & \textbf{$R^2$}           & \textbf{RMSE}            \\ \hline
                                                             & 2          & $0.666_{0.000}$          & $-0.235_{0.319}$         & $0.135_{0.073}$          &           & $0.742_{0.005}$                                  & $-0.198_{0.147}$                                 & $0.110_{0.059}$                                  &  & $1.000_{0.000}$          & $-11.524_{15.702}$       & $0.218_{0.105}$          \\
                                                             & 3          & $0.210_{0.063}$          & $0.073_{0.040}$          & $0.087_{0.012}$          &           & $0.772_{0.032}$                                  & $0.087_{0.026}$                                  & $0.081_{0.006}$                                  &  & $0.649_{0.012}$          & $0.032_{0.019}$          & $0.101_{0.009}$          \\
                                                             & 4          & $0.109_{0.005}$          & $0.173_{0.073}$          & $0.074_{0.006}$          &           & $0.701_{0.009}$                                  & $0.203_{0.024}$                                  & $0.079_{0.003}$                                  &  & $0.607_{0.005}$          & $0.078_{0.061}$          & $0.101_{0.015}$          \\
                                                             & 5          & $0.064_{0.011}$          & $0.187_{0.101}$          & $0.076_{0.006}$          &           & \cellcolor[HTML]{DFDFDF}\textbf{0.673$_{0.024}$} & \cellcolor[HTML]{DFDFDF}\textbf{0.345$_{0.066}$} & \cellcolor[HTML]{DFDFDF}\textbf{0.067$_{0.003}$} &  & $0.595_{0.008}$          & $0.209_{0.082}$          & $0.082_{0.016}$          \\
                                                             & 6          & $0.038_{0.006}$          & $0.052_{0.040}$          & $0.078_{0.005}$          &           & $0.628_{0.005}$                                  & $0.229_{0.057}$                                  & $0.060_{0.016}$                                  &  & $0.567_{0.004}$          & $0.126_{0.034}$          & $0.067_{0.000}$          \\
                                                             & 7          & $0.037_{0.001}$          & $0.242_{0.029}$          & $0.059_{0.007}$          &           & $0.597_{0.013}$                                  & $0.024_{0.242}$                                  & $0.068_{0.027}$                                  &  & $0.564_{0.004}$          & $0.182_{0.120}$          & $0.076_{0.012}$          \\
                                                             & 8          & \textbf{0.024$_{0.004}$} & \textbf{0.345$_{0.054}$} & \textbf{0.065$_{0.006}$} &           & $0.585_{0.022}$                                  & $0.126_{0.223}$                                  & $0.093_{0.028}$                                  &  & \textbf{0.563$_{0.003}$} & \textbf{0.252$_{0.084}$} & \textbf{0.066$_{0.007}$} \\
                                                             & 9          & $0.032_{0.003}$          & $0.157_{0.108}$          & $0.100_{0.006}$          &           & $0.619_{0.019}$                                  & $-0.376_{0.788}$                                 & $0.111_{0.036}$                                  &  & $0.552_{0.004}$          & $-0.213_{0.257}$         & $0.121_{0.055}$          \\
                                                             & 10         & $0.024_{0.004}$          & $0.233_{0.065}$          & $0.067_{0.005}$          &           & $0.592_{0.021}$                                  & $-1.090_{0.888}$                                 & $0.226_{0.075}$                                  &  & $0.569_{0.001}$          & $-0.343_{0.154}$         & $0.213_{0.065}$          \\
                                                             & 11         & $0.032_{0.003}$          & $0.267_{0.114}$          & $0.082_{0.016}$          &           & $0.587_{0.011}$                                  & $-4.903_{7.114}$                                 & $0.116_{0.052}$                                  &  & $0.566_{0.004}$          & $-0.135_{0.245}$         & $0.133_{0.076}$          \\
\multirow{-11}{*}{\textbf{\rotatebox{90}{Velocity anomaly}}} & 12         & $0.018_{0.002}$          & $0.233_{0.065}$          & $0.069_{0.008}$          &           & $0.589_{0.012}$                                  & $-7.192_{7.829}$                                 & $0.217_{0.026}$                                  &  & $0.573_{0.002}$          & $-0.270_{0.065}$         & $0.158_{0.046}$          \\ \hline
                                                             & 2          & $0.600_{0.000}$          & $-0.260_{0.248}$         & $0.137_{0.023}$          &           & $0.750_{0.000}$                                  & $-0.126_{0.298}$                                 & $0.128_{0.028}$                                  &  & $1.000_{0.000}$          & $-25.046_{1.777}$        & $0.394_{0.036}$          \\
                                                             & 3          & $0.219_{0.006}$          & $-0.321_{0.151}$         & $0.148_{0.022}$          &           & $0.689_{0.017}$                                  & $-0.153_{0.188}$                                 & $0.138_{0.014}$                                  &  & $0.641_{0.007}$          & $-0.488_{0.126}$         & $0.161_{0.039}$          \\
                                                             & 4          & $0.095_{0.005}$          & $0.144_{0.126}$          & $0.092_{0.010}$          &           & $0.683_{0.016}$                                  & $0.305_{0.049}$                                  & $0.092_{0.008}$                                  &  & $0.636_{0.009}$          & $0.196_{0.059}$          & $0.087_{0.016}$          \\
                                                             & 5          & $0.064_{0.005}$          & $0.296_{0.068}$          & $0.077_{0.005}$          &           & $0.721_{0.012}$                                  & $0.284_{0.014}$                                  & $0.092_{0.007}$                                  &  & $0.642_{0.005}$          & $-1.564_{2.523}$         & $0.224_{0.162}$          \\
                                                             & 6          & $0.064_{0.003}$          & $0.268_{0.019}$          & $0.087_{0.005}$          &           & \cellcolor[HTML]{DFDFDF}\textbf{0.676$_{0.019}$} & \cellcolor[HTML]{DFDFDF}\textbf{0.395$_{0.072}$} & \cellcolor[HTML]{DFDFDF}\textbf{0.088$_{0.008}$} &  & $0.634_{0.018}$          & $-0.866_{1.435}$         & $0.217_{0.153}$          \\
                                                             & 7          & \textbf{0.030$_{0.001}$} & \textbf{0.337$_{0.086}$} & \textbf{0.078$_{0.005}$} &           & $0.631_{0.008}$                                  & $0.314_{0.095}$                                  & $0.101_{0.020}$                                  &  & $0.592_{0.008}$          & $-1.901_{2.963}$         & $0.207_{0.173}$          \\
                                                             & 8          & $0.045_{0.004}$          & $0.219_{0.073}$          & $0.083_{0.004}$          &           & $0.669_{0.015}$                                  & $0.296_{0.079}$                                  & $0.092_{0.004}$                                  &  & \textbf{0.616$_{0.005}$} & \textbf{0.242$_{0.067}$} & \textbf{0.090$_{0.001}$} \\
                                                             & 9          & $0.037_{0.004}$          & $0.277_{0.072}$          & $0.070_{0.009}$          &           & $0.658_{0.011}$                                  & $0.296_{0.079}$                                  & $0.070_{0.004}$                                  &  & $0.654_{0.009}$          & $-1.803_{2.457}$         & $0.257_{0.146}$          \\
                                                             & 10         & $0.038_{0.007}$          & $0.312_{0.064}$          & $0.069_{0.007}$          &           & $0.674_{0.015}$                                  & $0.358_{0.100}$                                  & $0.084_{0.011}$                                  &  & $0.659_{0.019}$          & $-0.944_{1.696}$         & $0.203_{0.161}$          \\
                                                             & 11         & $0.017_{0.001}$          & $0.210_{0.024}$          & $0.096_{0.003}$          &           & $0.663_{0.012}$                                  & $0.124_{0.022}$                                  & $0.100_{0.009}$                                  &  & $0.641_{0.012}$          & $-2.743_{2.283}$         & $0.320_{0.144}$          \\
\multirow{-11}{*}{\textbf{\rotatebox{90}{Weight anomaly}}}   & 12         & $0.029_{0.006}$          & $0.171_{0.062}$          & $0.100_{0.009}$          &           & $0.664_{0.010}$                                  & $0.142_{0.031}$                                  & $0.090_{0.009}$                                  &  & $0.647_{0.007}$          & $-2.800_{0.425}$         & $0.415_{0.016}$          \\ \hline
\end{tabular}}
    \label{tab:modeling_results_road}
\end{table*}
\begin{figure*}[!t]
    \centering
    \includegraphics[width=\textwidth]{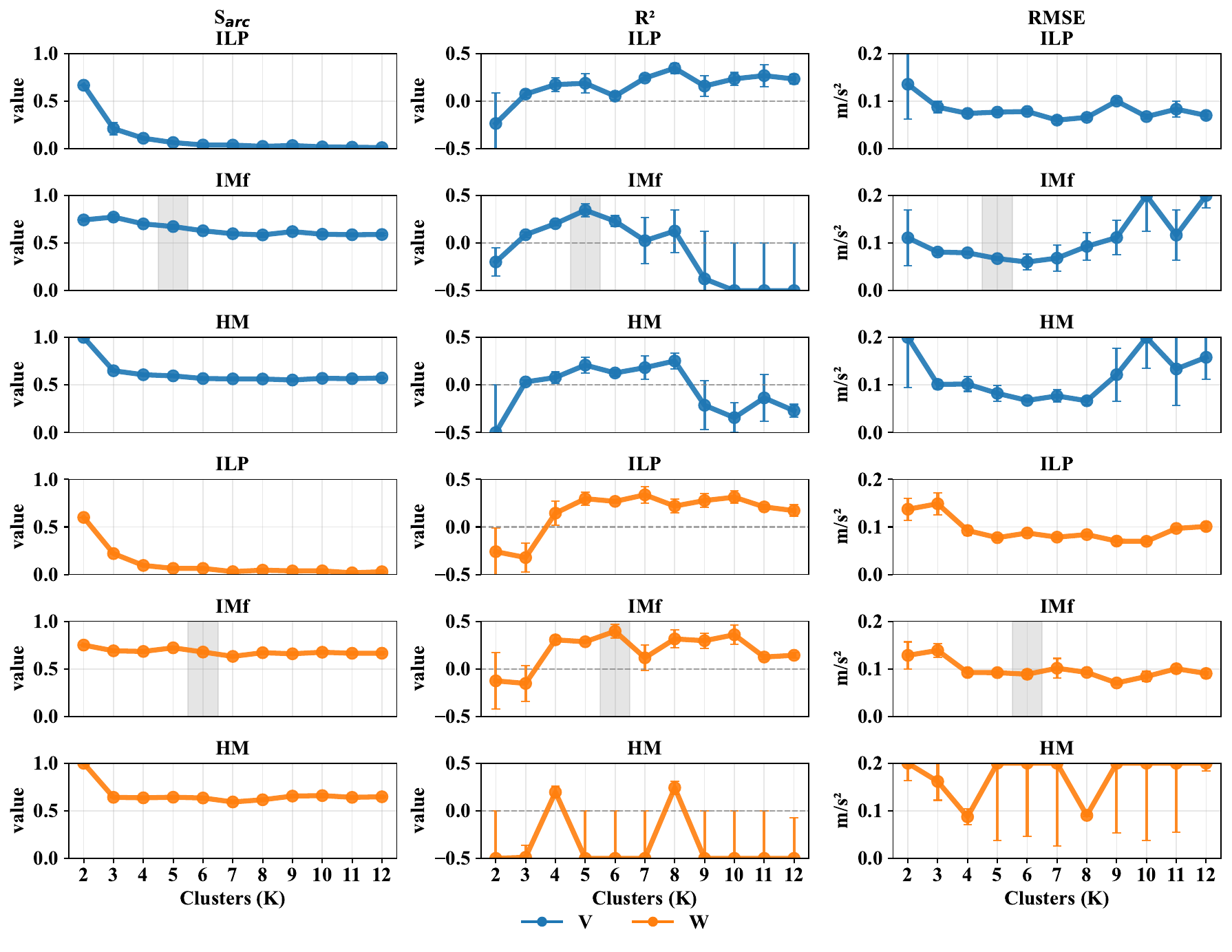}
    \caption{The S$_{arc}$, $R^2$ and RMSE trends for each process discovery algorithm with an increasing number of clusters K, for both the velocity (V) and weight (W) anomalies. The grey regions indicate the Petri nets achieving the best trade-off according to the three metrics.}
    \label{fig:plot_metrics}
\end{figure*}

\subsection{Modeling and simulation quality}
To assess the impact of clustering and process discovery configurations on modeling and simulation metrics, we outline a two-factor full factorial experiment, where the first factor is K, and the second factor is the process discovery algorithm. Table \ref{tab:modeling_results_road} reports the method results for the S$_{arc}$, R$^2$ and RMSE metrics. Additionally, Figure \ref{fig:plot_metrics} shows the S$_{arc}$, R$^2$, and RMSE trends for each process discovery algorithm with an increasing number of clusters K.

\subsubsection{Results}

\paragraph{S$_{arc}$ impact}
The interpretability of the Petri net in terms of S$_{arc}$ tends to drop for each process discovery algorithm as K increases. This trend is especially true for the \gls{ilp}, for which S$_{arc}$ drops abruptly starting from K=3 for both the velocity and weight anomalies. This trend is true for the \gls{imf} and HM too, although it tends to stabilize for both anomalies (S$_{arc}\geq$ 0.552). Such low S$_{arc}$ values for the \gls{ilp} are due to its tendency to produce overfitting Petri nets, as the algorithm attempts to compact too many behaviors in the same control-flow structure. In turn, the \gls{ilp} appears unsuitable for low-level sensor data due to the very low interpretability of its resulting Petri nets.

\paragraph{R$^2$ impact}
The impact on R$^2$ is especially evident for low K values, for which the simple state space is unable to capture the variability of the low-level sensor data. For example, considering the HM and weight anomaly for K=2, R$^2$ is equal to -25.046, which indicates a strong lack of variance explanation by the simulated version of the signal. This is also true for the \gls{ilp} and \gls{imf}, which reveal low R$^2$ values for K=2 (-0.235 and -0.198 for the velocity anomaly and -0.260 and -0.126 for the weight anomaly). However, R$^2$ tends to increase for all the process discovery algorithms. For example, the \gls{imf} peaks, for the velocity anomaly and weight anomaly, at 0.345 for K=5 and 0.395 for K=6, respectively. However, as the state space increases in complexity, R$^2$ tends to drop. This is true for all process discovery algorithms, although the \gls{ilp} appears to be more stable. Still, the \gls{imf} achieves the highest R$^2$ for both anomaly types (0.345 for K=5 and 0.395 for K=6).

\paragraph{RMSE impact}
RMSE values follow a trend very similar to R$^2$, although less visible due to the metric being concerned with amplitudes rather than variance. However, it is quite clear, especially from the \gls{imf} and HM, that the RMSE achieves a minimum value and then begins to rise again. For example, the RMSE of the \gls{imf}, for the velocity and weight anomaly, drops down to 0.060 for K=6 and 0.088 for K=6, and then rises up to 0.226 for K=10 and 0.100 for K=11. Again, the \gls{ilp} appears to be more stable than the other two; however, the \gls{imf} is the most balanced, which confirms that it achieves the best trade-off among the three metrics.

\subsubsection{Interpretation and root cause analysis} 
\begin{figure*}[p]
    \centering
    \includegraphics[width=.95\textwidth]{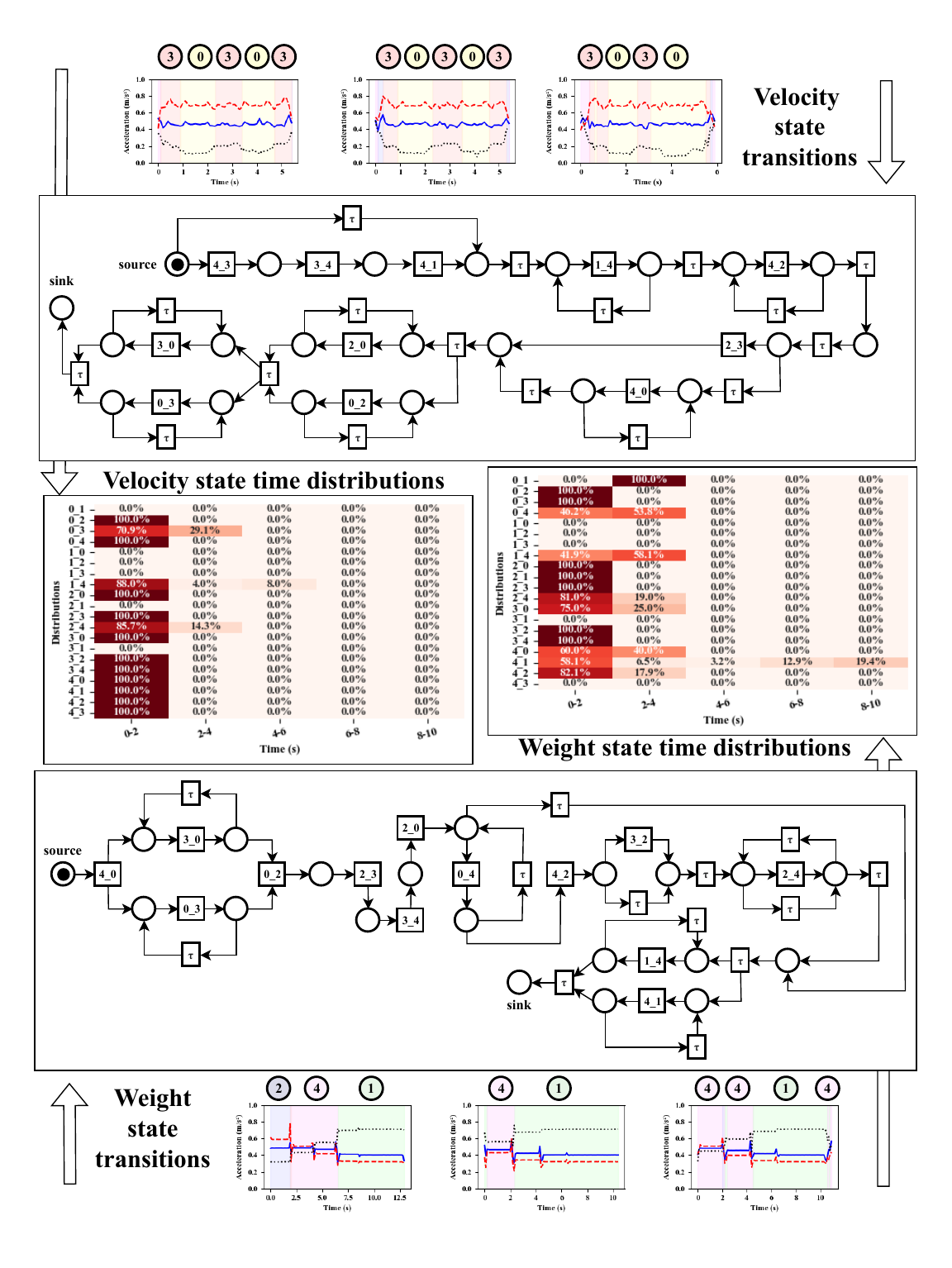}
    \caption{The control-flow structure and time perspective of the velocity and weight Petri nets generated through our method with the \gls{imf} and K=5.}
    \label{fig:interpretation}
\end{figure*}

The Petri nets built by the process discovery algorithms can be linked to the physical properties of the \gls{cps} thanks to the characterization of the state transitions (Definition \ref{def:state_transition}) and state time distributions (Definition \ref{def:state_distribution}) of our method. Figure \ref{fig:interpretation} shows the control-flow structure and time perspective of the velocity (top) and weight (bottom) Petri nets built by the \gls{imf} with K=5. The velocity windows at the top and weight windows at the bottom have been clustered into regions (the colored segments in the windows' plots) in the event log extraction phase of the method. Each cluster $s$ represents a triple $(x,y,z)$, where the three elements are the centroid values of the robotic arm's acceleration at the selected joint. Therefore, each state transition can be linked to \gls{cps} dynamics, and replicated through the simulation phase of our method, generating simulated anomalous time series windows (Definition \ref{def:sim_ts}).

Each state transition recorded in the event log contributes to building the control-flow structure, which is different for the two types of anomalies. A clear difference is the looping of the 3 $\rightarrow$ 0 and 0 $\rightarrow$ 3 state transitions. This is consistent with the sample velocity time series windows at the top, in which, after a negligible alternation of state transitions of little duration, these two state transitions appear repeatedly. While the weight anomaly also has this looping, it only appears at the start, which then follows with the main loop, which involves the 4 $\rightarrow$ 1 and 1 $\rightarrow$ 4 state transitions. This can be observed from the weight time series windows at the bottom.
There are also state transitions that only appear exclusively in either the velocity or weight Petri net, such as the 4 $\rightarrow$ 3 state transition. It is worth noting that the Petri nets are influenced by the rapid state transitions that appear at the start of the time series window, which suggests that these should be discarded prior to their generation.

The time perspective obtained by the state time distributions of the velocity and weight time series windows also reveals insightful information about the root causes of each anomaly. For example, the state time distribution heatmap of the weight anomaly reveals that some state times associated with specific state transitions can last for a long time with a fairly high probability, such as the 4 $\rightarrow$ 1 state transition, whose duration can be up to the 8-10 second interval with 19.4\% probability. On the other hand, the time perspective of the velocity anomaly Petri net reveals that there are more state transitions occurring with shorter durations, while some state transitions are lengthier (within the 2-4 second range) with higher probabilities.

\subsection{Fault identification} 
\begin{table*}[!t]
    \centering
    \caption{The fault identification results in terms of F1 score and conformance checking (CC) time of our fault diagnosis method for both the velocity and weight anomaly. |P|+|Tr| indicates the total number of nodes in the Petri net extracted with the given process discovery algorithm–number of clusters combination, where |P| and |Tr| denote the cardinality of the sets of places and transitions, respectively. \#E indicates the number of events in the event logs built with a given number of clusters. TE indicates that conformance checking exceeded the 300-second computation time, whereas NS indicates that the Petri net obtained with the process discovery algorithm-number of clusters combination is non-sound. The grey cells highlight the results of the best Petri nets identified in the modeling and simulation quality experiment. Bold numbers highlight the best F1 per process discovery algorithm and anomaly type.}
    \label{tab:fault_classification}
    \resizebox{\textwidth}{!}{\begin{tabular}{clllllllllllll}
\hline
\multicolumn{1}{l}{}                                         & \textbf{}  & \multicolumn{3}{l}{\textbf{ILP (75\% noise tolerance)}}                                &  & \multicolumn{3}{l}{\textbf{IMF (75\% noise tolerance)}}                                                                & \textbf{} & \multicolumn{3}{l}{\textbf{HM (75\% noise tolerance)}}                                &                                \\ \cline{3-5} \cline{7-9} \cline{11-13}
\multicolumn{1}{l}{\textbf{}}                                & \textbf{K} & \textbf{F1 (\%)}               & \textbf{CC time (s)}         & |P|+|Tr|               &  & \textbf{F1 (\%)}                          & \textbf{CC time (s)}                    & |P|+|Tr|                         &           & \textbf{F1 (\%)}              & \textbf{CC time (s)}         & |P|+|Tr|               & \multirow{-2}{*}{\textbf{\#E}} \\ \cline{2-5} \cline{7-9} \cline{11-14} 
                                                             & 2          & $32.937_{27.222}$              & $0.004_{0.000}$              & $8_{0}$                &  & $20.392_{17.306}$                         & $0.004_{0.000}$                         & $15_{3}$                         &           & $65.627_{13.064}$             & $0.003_{0.000}$              & $4_{0}$                & $67_{4}$                       \\
                                                             & 3          & $60.103_{13.602}$              & $0.006_{0.000}$              & $15_{3}$               &  & $77.520_{4.769}$                          & $0.007_{0.000}$                         & $22_{3}$                         &           & $62.553_{17.117}$             & $0.007_{0.000}$              & $25_{1}$               & $156_{15}$                     \\
                                                             & 4          & $99.471_{0.748}$               & $0.011_{0.000}$              & $22_{2}$               &  & $\boldsymbol{98.925_{1.521}}$             & $\boldsymbol{0.010_{0.000}}$            & $\boldsymbol{39_{6}}$            &           & $86.734_{12.324}$             & $0.008_{0.001}$              & $44_{1}$               & $209_{30}$                     \\
                                                             & 5          & $96.736_{2.723}$               & $0.026_{0.003}$              & $36_{9}$               &  & \cellcolor[HTML]{DFDFDF}$93.504_{7.093}$  & \cellcolor[HTML]{DFDFDF}$0.020_{0.002}$ & \cellcolor[HTML]{DFDFDF}$61_{7}$ &           & $79.976_{2.783}$              & $0.015_{0.004}$              & $68_{3}$               & $300_{10}$                     \\
                                                             & 6          & $97.980_{2.857}$               & $0.061_{0.019}$              & $60_{12}$              &  & $97.072_{3.117}$                          & $0.158_{0.091}$                         & $80_{9}$                         &           & $\boldsymbol{99.206_{0.794}}$ & $\boldsymbol{0.029_{0.013}}$ & $\boldsymbol{102_{3}}$ & $480_{18}$                     \\
                                                             & 7          & $\boldsymbol{100.000_{0.000}}$ & $\boldsymbol{0.129_{0.058}}$ & $\boldsymbol{58_{10}}$ &  & $81.400_{12.258}$                         & $1.478_{1.369}$                         & $81_{13}$                        &           & $92.649_{4.125}$              & $0.029_{0.001}$              & $118_{1}$              & $537_{12}$                     \\
                                                             & 8          & $99.487_{0.725}$               & $0.177_{0.086}$              & $80_{15}$              &  & $73.462_{22.963}$                         & $2.303_{1.553}$                         & $97_{8}$                         &           & $91.429_{0.000}$              & $0.054_{0.000}$              & $146_{2}$              & $563_{16}$                     \\
                                                             & 9          & $100.000_{0.000}$              & $0.145_{0.032}$              & $107_{1}$              &  & $70.720_{34.408}$                         & $1.205_{0.966}$                         & $111_{3}$                        &           & NS                            & NS                           & $187_{8}$              & $614_{14}$                     \\
                                                             & 10         & $100.000_{0.000}$              & $0.239_{0.030}$              & $125_{6}$              &  & $21.622_{0.000}$                          & $8.899_{0.000}$                         & $180_{20}$                       &           & NS                            & NS                           & $188_{8}$              & $733_{10}$                     \\
                                                             & 11         & $100.000_{0.000}$              & $0.507_{0.215}$              & $131_{24}$             &  & $81.871_{11.311}$                         & $64.984_{53.965}$                       & $178_{33}$                       &           & NS                            & NS                           & $180_{12}$             & $763_{17}$                     \\
\multirow{-11}{*}{\textbf{\rotatebox{90}{Velocity anomaly}}} & 12         & $100.000_{0.000}$              & $1.491_{1.501}$              & $156_{3}$              &  & TE                                        & TE                                      & $246_{24}$                       &           & NS                            & NS                           & $206_{18}$             & $763_{25}$                     \\ \hline
                                                             & 2          & $15.528_{8.305}$               & $0.004_{0.000}$              & $9_{0}$                &  & $12.472_{14.845}$                         & $0.005_{0.000}$                         & $12_{0}$                         &           & $0.000_{0.000}$               & $0.003_{0.000}$              & $4_{0}$                & $48_{3}$                       \\
                                                             & 3          & $61.187_{9.589}$               & $0.007_{0.000}$              & $15_{1}$               &  & $73.447_{3.501}$                          & $0.007_{0.000}$                         & $21_{1}$                         &           & $49.313_{9.275}$              & $0.006_{0.000}$              & $29_{1}$               & $68_{6}$                       \\
                                                             & 4          & $98.990_{1.429}$               & $0.013_{0.002}$              & $23_{1}$               &  & $\boldsymbol{98.039_{2.773}}$             & $\boldsymbol{0.015_{0.005}}$            & $\boldsymbol{31_{4}}$            &           & $82.185_{9.047}$              & $0.013_{0.003}$              & $38_{4}$               & $118_{15}$                     \\
                                                             & 5          & $94.213_{4.548}$               & $0.027_{0.007}$              & $30_{2}$               &  & $90.600_{9.234}$                          & $0.032_{0.003}$                         & $43_{4}$                         &           & $70.175_{3.509}$              & $0.017_{0.006}$              & $43_{11}$              & $114_{19}$                     \\
                                                             & 6          & $95.556_{6.285}$               & $0.045_{0.016}$              & $30_{1}$               &  & \cellcolor[HTML]{DFDFDF}$91.232_{10.204}$ & \cellcolor[HTML]{DFDFDF}$0.061_{0.032}$ & \cellcolor[HTML]{DFDFDF}$60_{9}$ &           & $\boldsymbol{98.485_{1.515}}$ & $\boldsymbol{0.037_{0.021}}$ & $\boldsymbol{43_{14}}$ & $139_{37}$                     \\
                                                             & 7          & $\boldsymbol{100.000_{0.000}}$ & $\boldsymbol{0.037_{0.006}}$ & $\boldsymbol{61_{7}}$  &  & $74.100_{13.199}$                         & $0.119_{0.063}$                         & $103_{8}$                        &           & $87.059_{7.059}$              & $0.018_{0.001}$              & $58_{6}$               & $119_{3}$                      \\
                                                             & 8          & $98.925_{1.521}$               & $1.731_{2.255}$              & $48_{1}$               &  & $70.381_{21.026}$                         & $0.309_{0.056}$                         & $114_{22}$                       &           & $76.923_{0.000}$              & $0.105_{0.000}$              & $57_{6}$               & $148_{15}$                     \\
                                                             & 9          & $100.000_{0.000}$              & $0.096_{0.026}$              & $56_{3}$               &  & $78.930_{18.599}$                         & $0.625_{0.393}$                         & $123_{16}$                       &           & NS                            & NS                           & $60_{10}$              & $180_{28}$                     \\
                                                             & 10         & $100.000_{0.000}$              & $0.160_{0.032}$              & $60_{3}$               &  & $50.848_{0.000}$                          & $16.307_{0.000}$                        & $119_{27}$                       &           & NS                            & NS                           & $41_{7}$               & $158_{44}$                     \\
                                                             & 11         & $100.000_{0.000}$              & $1.309_{1.639}$              & $92_{6}$               &  & $65.662_{24.284}$                         & $0.519_{0.451}$                         & $123_{22}$                       &           & NS                            & NS                           & $83_{4}$               & $160_{38}$                     \\
\multirow{-11}{*}{\textbf{\rotatebox{90}{Weight anomaly}}}   & 12         & $100.000_{0.000}$              & $0.262_{0.051}$              & $136_{25}$             &  & TE                                        & TE                                      & $192_{61}$                       &           & NS                            & NS                           & $72_{12}$              & $202_{14}$                     \\ \hline
\end{tabular}}
\end{table*}
\begin{figure*}[!t]
\centering
    \includegraphics[width=\textwidth]{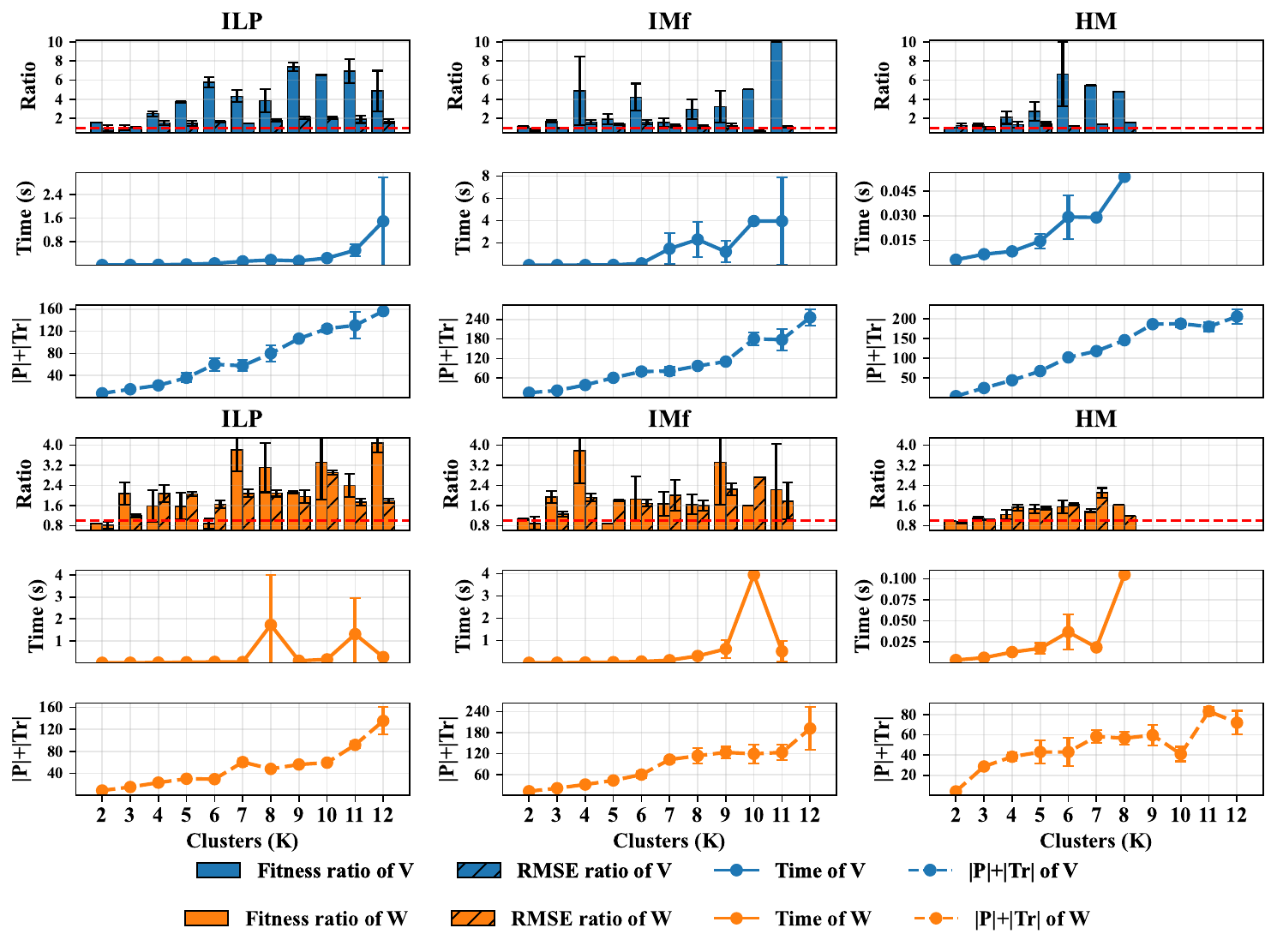}
    \caption{The fitness and RMSE ratio for each process discovery algorithm with an increasing number of clusters K, for both the velocity (V) and weight (W) anomalies. Below each ratio is the conformance checking time and the structural complexity of the Petri nets in terms of the number of places and transitions (|P|+|Tr|).}
    \label{fig:fault_identification}
\end{figure*}

To assess the impact of clustering and process discovery configurations on the ability of our method to correctly distinguish faults from one another, we outline a two-factor full factorial experiment, where the first factor is K, and the second factor is the process discovery algorithm. Table \ref{tab:fault_classification} reports the method results for the F1 and conformance checking time metrics. The N/A values indicate that conformance checking could not be performed due to the Petri net being non-sound or because the computation exceeded the total time limit of 300 seconds. 
Additionally, Figure \ref{fig:fault_identification} provides an interpretation of the results according to the ratios achieved by the fitness and RMSE metrics between positives and negatives, which play a pivotal role in fault identification (see Algorithm \ref{fi_algo}).

\subsubsection{Results}

\paragraph{F1 impact} The classification ability of the Petri nets depends heavily on the number of clusters K. For example, considering the weight anomaly, the \gls{ilp}, \gls{imf}, and HM have as low as 15.528\%, 12.472\%, and 0.000\% F1 values. On the other hand, for higher K values, F1 rises, with the \gls{ilp} achieving perfect performance with K=7 for both the velocity and weight anomalies. Similarly, considering the velocity anomaly, the \gls{imf} and HM achieve F1 scores of 98.925\% and 96.630\% for K=4 and K=8, respectively. However, while the \gls{ilp} maintains perfect classification for K$\geq$7, the \gls{imf} and HM performance drop, which can be attributed to the results shown in the previous experiment. Notably, the \gls{ilp} is unable to classify any trace for K=8, 9, and 10. This is due to the Petri net being non-sound, which makes it unsuitable for performing conformance checking. On the other hand, the \gls{imf} always produces sound Petri nets, so this process discovery algorithm is safe in this regard. These results are backed up by the visual interpretation provided in Figure \ref{fig:fault_identification}, which shows that with increasing numbers of clusters, the ratios between positives and negatives of two out of the three metrics used during fault identification get higher than 1, meaning that there is a clear distinction between the anomalous velocity and weight time series windows.

\paragraph{Time impact} The conformance checking time tends to increase for each algorithm as the state complexity (K) of the Petri net increases. This trend is especially visible in the \gls{imf}, which achieves velocities and weights anomalies as low as 0.003 and 0.004 seconds at K=2, and peaks at 2.363 and 0.691 seconds at K=8 and 9, respectively. It finally exceeds the set time limit for K=10, 11, and 12. This is due to both the size of the Petri net in terms of the number of places and arcs, and the number of events of the event logs to check. In fact, the total number of places and transitions (|P|+|Tr|) of the Petri nets and the number of events (\#E) increase significantly as the number of clusters increases, leading to increasing conformance checking time too, as shown in Figure \ref{fig:fault_identification}. Such behavior is to be expected, as conformance checking time tends to worsen as the number of events and structural complexity of the Petri net increase \cite{aalst2022pmhb}. At times, the behavior of the conformance checking time is quite erratic. In fact, regarding the differences across various process discovery algorithms, the \gls{imf} shows some unexpected peaks, such as 16.307 seconds for the weight anomaly at K=10. It appears that \gls{ilp} and HM are more stable, although the former has very low simplicity, whereas the latter may discover non-sound (NS) Petri nets.

\subsubsection{Ablation test and comparisons}
\begin{table}[!t]
\centering
\resizebox{\columnwidth}{!}{{\begin{tabular}{clllll}
\hline
\textbf{}                                                   & \textbf{Acc} & \textbf{Ours}                        & \textbf{LSTM-AE}                     & \textbf{BiLSTM-AE}                    & \textbf{GRU-AE}                       \\ \hline
\multirow{10}{*}{\textbf{\rotatebox{90}{Velocity anomaly}}} & 100\%          & \textbf{98.430\textsubscript{0.020}} & 89.700\textsubscript{7.830}          & 94.550\textsubscript{1.570}           & 92.810\textsubscript{2.110}           \\
                                                            & 90\%           & 82.170\textsubscript{4.570}          & \textbf{87.230\textsubscript{6.130}} & 87.120\textsubscript{3.680}           & 86.290\textsubscript{0.750}           \\
                                                            & 80\%           & \textbf{84.580\textsubscript{2.420}} & 82.160\textsubscript{3.610}          & 82.630\textsubscript{1.700}           & 83.210\textsubscript{6.220}           \\
                                                            & 70\%           & 57.710\textsubscript{14.210}         & 82.810\textsubscript{3.320}          & \textbf{84.120\textsubscript{4.390}}  & 82.490\textsubscript{3.010}           \\
                                                            & 60\%           & 39.750\textsubscript{6.600}          & 81.540\textsubscript{4.150}          & \textbf{83.610\textsubscript{6.500}}  & 79.950\textsubscript{1.560}           \\
                                                            & 50\%           & 48.380\textsubscript{12.160}         & 74.210\textsubscript{6.900}          & 77.430\textsubscript{6.670}           & \textbf{79.350\textsubscript{7.850}}  \\
                                                            & 40\%           & 41.940\textsubscript{22.020}         & 64.140\textsubscript{11.480}         & \textbf{67.060\textsubscript{19.770}} & 61.220\textsubscript{14.050}          \\
                                                            & 30\%           & 25.560\textsubscript{9.890}          & 58.400\textsubscript{2.300}          & \textbf{70.130\textsubscript{6.550}}  & 66.230\textsubscript{13.920}          \\
                                                            & 20\%           & 24.540\textsubscript{15.810}         & 48.660\textsubscript{5.630}          & 40.870\textsubscript{5.190}           & \textbf{58.950\textsubscript{22.850}} \\
                                                            & 10\%           & 11.700\textsubscript{9.750}          & 20.570\textsubscript{11.060}         & 16.880\textsubscript{4.270}           & \textbf{37.210\textsubscript{19.840}} \\ \hline
\multirow{10}{*}{\textbf{\rotatebox{90}{Weight anomaly}}}   & 100\%          & \textbf{96.900\textsubscript{0.090}} & 84.150\textsubscript{3.360}          & 84.270\textsubscript{8.980}           & 83.170\textsubscript{11.670}          \\
                                                            & 90\%           & 58.150\textsubscript{15.530}         & 75.540\textsubscript{4.890}          & \textbf{77.510\textsubscript{5.700}}  & 67.810\textsubscript{3.630}           \\
                                                            & 80\%           & 60.680\textsubscript{17.170}         & \textbf{64.020\textsubscript{1.050}} & 61.470\textsubscript{0.870}           & 57.950\textsubscript{2.660}           \\
                                                            & 70\%           & 39.930\textsubscript{18.960}         & \textbf{61.960\textsubscript{1.120}} & 54.430\textsubscript{7.190}           & 48.140\textsubscript{3.340}           \\
                                                            & 60\%          & 45.580\textsubscript{9.300}          & \textbf{58.340\textsubscript{4.380}} & 50.000\textsubscript{6.630}           & 41.380\textsubscript{6.940}           \\
                                                            & 50\%           & 36.240\textsubscript{26.770}         & \textbf{54.910\textsubscript{4.030}} & 51.080\textsubscript{0.370}           & 46.740\textsubscript{5.790}           \\
                                                            & 40\%           & 45.160\textsubscript{16.340}         & \textbf{53.800\textsubscript{2.400}} & 45.620\textsubscript{1.760}           & 46.850\textsubscript{6.790}           \\
                                                            & 30\%           & 12.000\textsubscript{16.970}         & \textbf{50.830\textsubscript{0.030}} & 41.190\textsubscript{4.380}           & 38.280\textsubscript{10.510}          \\
                                                            & 20\%           & 25.770\textsubscript{17.920}         & \textbf{49.800\textsubscript{5.280}} & 43.440\textsubscript{6.420}           & 34.930\textsubscript{11.340}          \\
                                                            & 10\%           & 14.440\textsubscript{20.430}         & 20.570\textsubscript{11.060}         & 16.880\textsubscript{4.270}           & \textbf{37.210\textsubscript{19.840}} \\ \hline
\end{tabular}%
}}
\caption{Fault identification comparison between our method and other deep learning-based approaches with decreasing anomaly detection accuracy (Acc) percentages. Bold figures indicate the best F1 per Acc percentage.}
\label{tab:ablation_test}
\end{table}
To further test the F1 performance of our method, we compared our method with other deep learning-based approaches (GRU-AE \cite{guo2018gru}, LSTM-AE \cite{malhotra2016lstm}, and BiLSTM-AE \cite{raihan2023bi}) with decreasing percentages of Acc (from 100\% to 10\%, see Equation \ref{eq:acc}) to evaluate the impact of this phase on the subsequent results. The chosen deep learning-based approaches can be adapted to the one-class classification setting of this study, where the training set only contains the sample of the target anomaly. For the comparison, we have used the \gls{imf} with K=5.

\begin{figure}[!t]
\centering
    \includegraphics[width=\columnwidth]{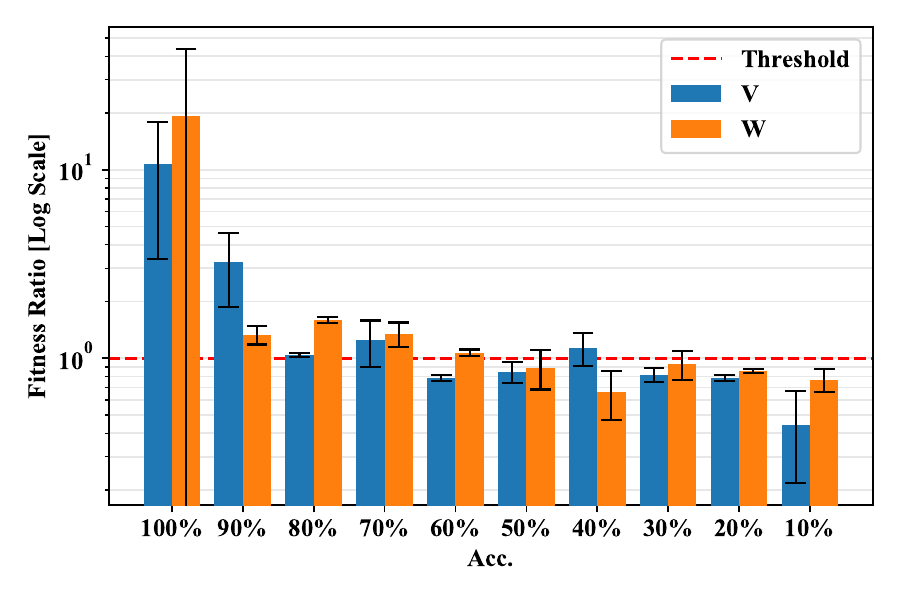}
    \caption{The fitness ratios for both the velocity (V) and weight (W) anomalies achieved by the \gls{imf} with K=5 per Acc percentage.}
    \label{fig:ratios}
\end{figure}

The F1 results are shown in Table \ref{tab:ablation_test}. In the scenario where anomaly detection is flawless (Acc=100\%), our method outperforms the other one-class approaches. However, as Acc values decrease, the other methods tend to perform better, although all of them drop to very low percentages when Acc reaches low values. This is to be expected: when the training set for fault diagnosis is contaminated with normal time series windows, all the models are unable to properly characterize the specific fault, leading to worse performance. It is important to note that our method using the \gls{imf} with K=5 is less robust to contamination in normal behavior. This behavior can be explained using Figure \ref{fig:ratios}. Higher accuracy (Acc) leads to higher fitness ratios between positives and negatives, which in turn results in better fault identification. As the accuracy decreases, the fitness ratio also declines, along with the quality of fault identification. This effect is much more pronounced for our proposed method than for the deep learning approaches. While this may be partly due to the sub-optimal choice of the \gls{imf} with K=5, it may also indicate that deep learning approaches are inherently more robust to noise~\cite{wang2024dataperturbations}. However, the interpretation of deep learning methods is less straightforward because of their black-box nature.

\subsection{Discussion}
\paragraph{Modeling and simulation quality} In this part of the experimentation, we evaluated whether the methodology allowed the extraction of interpretable and accurate Petri nets. The experiments outlined that increasing the K factor led to slightly lower Petri net interpretability yet more accurate simulations in terms of R$^2$ and \gls{rmse}. However, this increase stops when the state space becomes too complex, causing the R$^2$ and \gls{rmse} to rise. Notably, while the \gls{ilp} seems the most stable process discovery algorithm in terms of simulation accuracy, its arc-degree simplicity (S$_{arc}$) drops drastically, making the Petri nets unreadable. On the other hand, the \gls{imf} is the most balanced algorithm, achieving good interpretability and surpassing the HM in simulation accuracy. In conclusion, the different process discovery algorithms significantly impact the interpretability of Petri nets, while a trade-off for state complexity must be found regardless of the algorithm used.

\paragraph{Fault Identification} Next, we evaluated whether the method enabled distinguishing between the two different anomaly types using the proposed fault identification algorithm. The K factor highly influences the F1-score results of our method, as low K values lead to poor classification results for all the process discovery algorithms and in both anomaly types. High K values may lead to higher F1-score, although it has been observed that this trend is not met by the \gls{imf}, whose performance tends to drop as K increases. Yet, while the \gls{ilp} meets this trend, the algorithm may build non-sound Petri nets, which hinders the ability to perform alignment-based conformance checking. Regarding the time needed, conformance checking becomes more computationally expensive as K increases. Such an increase is due to both the heightened number of places and arcs of the Petri nets and the number of events resulting from the increasing state space complexity. Finally, the ablation test and comparison highlighted that the accuracy of anomaly detection prior to fault identification has a significant impact on the F1 performance across our method and other deep learning-based approaches.

\subsection{Threats to validity}
The threats to validity regard possible construct, internal, and external pitfalls in the experimentation.

\paragraph{Construct validity} The experiments evaluated our method based on three aspects: Petri net interpretability, simulation accuracy, and fault diagnosis effectiveness. Petri net interpretability was measured via arc-degree simplicity, which provides information about structural properties but does not capture other control-flow aspects, such as cyclomatic complexity \cite{lassen2009complexitymetrics}. Simulation accuracy was assessed using RMSE and R$^2$, which reflect the similarity in amplitude and the variance explained by the simulated data, respectively. Other metrics, such as the Pearson correlation coefficient \cite{schober2018correlation}, could reveal additional aspects of simulation quality. Fault diagnosis effectiveness was evaluated using the F1-score, a standard classification metric; however, alternative metrics, such as the Matthews correlation coefficient~\cite{chicco2021matthews}, may provide complementary insights. Overall, while these metrics provide a valid and interpretable approximation of the aspects of interest, they may not fully capture all dimensions of process behavior.

\paragraph{Internal validity} We observed correlations between the state space complexity of the Petri net and the chosen process discovery algorithms on the modeling and simulation quality metrics, as shown in Table \ref{tab:modeling_results_road} and Figure \ref{fig:plot_metrics}. Similarly, these two factors have been shown to have a significant impact on fault identification capabilities, as illustrated in Table \ref{tab:fault_classification} and Figure \ref{fig:fault_identification}. Finally, we have also evaluated the impact of anomaly detection accuracy on fault identification, and compared the results to other state-of-the-art methods for time series classification, as shown in Table \ref{tab:ablation_test}. However, other factors may also have influenced these metrics. For example, the noisy nature of low-level sensor data can significantly impact the quality of the Petri nets. While we partially addressed this by setting a high noise threshold for all the algorithms, this challenge requires a more in-depth study.

\paragraph{External validity} Our study focused on the RoAD case study, which involved real multivariate time series data collected from various sensors mounted on a Kuka LBR iiwa robotic arm. While this dataset provides a solid foundation for evaluation, applying our methodology to additional \glspl{cps} would further strengthen the generalizability of our findings. Future studies could validate these results across different \gls{cps} environments to assess their broader applicability. Additionally, other faulty scenarios where regular trends do not follow each other as shown in Figure \ref{fig:ice_lab} should be studied to validate our approach to other manufacturing contexts.
\section{Conclusions}
\label{sec:conclusions}
The ever-increasing complexity of \glspl{cps} in production environments heightens the possibility of introducing vulnerabilities in their development and the occurrence of run-time faults. Therefore, reliable operation hinges on techniques that support interpretable and autonomous fault diagnosis capable of identifying deviations from normal behavior, classifying faults, and enhancing root-cause analysis. Yet, existing deep learning-based approaches yield black-box diagnoses that lack interpretability and the ability to understand the root causes of the occurred fault. On the other hand, Petri net-based approaches, valued for their clear, process-based semantics, either require manual, error-prone modeling by domain experts or do not directly target data-driven modeling of faulty \gls{cps} behaviors from low-level sensor data. 

This study has demonstrated that process mining, specifically the discovery of interpretable Petri nets enriched with stochastic timing information, offers a principled, data-driven alternative that can help address the outlined challenges. To this aim, we introduced a process mining-driven fault diagnosis method that isolates anomalous sensor windows, converts them into event logs, and automatically derives stochastic Petri nets capable of replicating and explaining faulty dynamics. Extensive experiments have been performed on the Robotic Arm Dataset (RoAD), a benchmark collected from a robotic arm deployed in a scale-replica smart manufacturing assembly line. Results outline that the Petri nets obtained through our method (i) are able to strike a balance between interpretability and simulation accuracy, (ii) identify different types of faulty reliably with low classification times. Collectively, these findings confirm the viability of the obtained Petri nets as transparent, executable knowledge sources for \gls{cps} dependability analyses.

Future research will focus on enhancing manufacturing system reliability by applying the proposed method across various industrial settings, including advanced manufacturing lines and supply chain networks. Integrating the discovered fault models with predictive maintenance and smart scheduling platforms can drive early fault identification, reduce unplanned downtime, and optimize resource allocation on the factory floor. Additionally, coupling simulation-based fault diagnosis with safety assurance tools, such as digital twins and runtime monitoring, can offer manufacturers quantitative insights into system robustness, enabling more informed decisions to maintain continuous, safe, and efficient production operations under fault conditions.

\section*{Declaration of competing interest}
The authors declare that they have no known competing financial interests or personal relationships that could have appeared to influence the work reported in this paper.

\section*{Acknowledgments}
This study is supported by the Spoke 9 “Digital Society \& Smart Cities” of ICSC - Centro Nazionale di Ricerca in High Performance-Computing, Big Data and Quantum Computing, funded by the European Union - NextGenerationEU (PNRR-HPC, CUP: E63C22000980007), by the European Union’s Horizon Europe research and innovation programme under the Marie Skłodowska-Curie grant agreement No. 101109243, and by the Veneto Region within the PR Veneto FESR 2021-2027 program, Action 1.1.1 Sub A ``Rafforzare la ricerca e l’innovazione tra imprese e organismi di ricerca'' (DGR n. 729/2024), project ``SUPREME'' (CUP D19J24000680007 – ID 24729\_001692). This manuscript reflects only the Authors’ views and opinions, neither the European Union nor the European Commission can be considered responsible for them.

\section*{Data availability}
The code and dataset for this paper are available in the following GitHub repository: \url{https://github.com/francescovitale/pm\_based\_modeling\_simulation}.

\bibliographystyle{elsarticle-num}
\bibliography{bibliography}

\end{document}